\documentclass[prodmode,acmcsur,letterpaper]{acmsmall}
\usepackage[ruled]{algorithm2e}
\usepackage{graphics}
\usepackage{caption}
\usepackage{subfigure}
\usepackage{url}
\usepackage{amssymb,amsmath}
\usepackage{color}
\usepackage{multirow}
\usepackage{rotating}
\usepackage{tikz}
\usepackage{hyperref}
\graphicspath{{./figure/}}

\newcommand{\mat}[1]{{\bf #1}}   
\newcommand{\argmin}{\arg\!\min}

\SetAlFnt{\small}
\SetAlCapFnt{\small}
\SetAlCapNameFnt{\small}
\SetAlCapHSkip{0pt}
\IncMargin{-\parindent}

\acmVolume{9}
\acmNumber{4}
\acmArticle{39}
\acmYear{2010}
\acmMonth{3}

\begin{document}

\markboth{J. Li et al.}{Feature Selection: A Data Perspective}
\title{Feature Selection: A Data Perspective}
\author{Jundong Li
\affil{Arizona State University}
Kewei Cheng
\affil{Arizona State University}
Suhang Wang
\affil{Arizona State University}
Fred Morstatter
\affil{Arizona State University}
Robert P. Trevino
\affil{Arizona State University}
Jiliang Tang
\affil{Michigan State University}
Huan Liu
\affil{Arizona State University}}

\begin{abstract}
Feature selection, as a data preprocessing strategy, has been proven to be effective and efficient in preparing data (especially high-dimensional data) for various data mining and machine learning problems. The objectives of feature selection include: building simpler and more comprehensible models, improving data mining performance, and preparing clean, understandable data. The recent proliferation of big data has presented some substantial challenges and opportunities to feature selection. In this survey, we provide a comprehensive and structured overview of recent advances in feature selection research. Motivated by current challenges and opportunities in the era of big data, we revisit feature selection research from a data perspective and review representative feature selection algorithms for conventional data, structured data, heterogeneous data and streaming data. Methodologically, to emphasize the differences and similarities of most existing feature selection algorithms for conventional data, we categorize them into four main groups: similarity based, information theoretical based, sparse learning based and statistical based methods. To facilitate and promote the research in this community, we also present an open-source feature selection repository that consists of most of the popular feature selection algorithms (\url{http://featureselection.asu.edu/}). Also, we use it as an example to show how to evaluate feature selection algorithms. At the end of the survey, we present a discussion about some open problems and challenges that require more attention in future research.
\end{abstract}



\acmformat{Jundong Li, Kewei Cheng, Suhang Wang, Fred Morstatter, Robert P. Trevino, Jiliang Tang,
and Huan Liu, 2017. Feature Selection: A Data Perspective}

\begin{bottomstuff}
This material is based upon work supported by, or in part by, the NSF grant 1217466 and 1614576.

Author's addresses: J. Li, K. Cheng, S. Wang, F. Morstatter, R.P. Trevino, H. Liu, Computer Science and Engineering, Arizona State University, Tempe, AZ, 85281;
email: \{jundongl, kcheng18, swang187, fmorstat, rptrevin, huan.liu\}@asu.edu; J. Tang, Michigan State University, East Lansing, MI 48824; email: tangjili@msu.edu.
\end{bottomstuff}

\maketitle

\vspace{-0.1in}
\section{Introduction}
We are now in the era of big data, where huge amounts of high-dimensional data become ubiquitous in a variety of domains, such as social media, healthcare, bioinformatics and online education. The rapid growth of data presents challenges for effective and efficient data management. It is desirable to apply data mining and machine learning techniques to automatically discover knowledge from data of various sorts.

When data mining and machine learning algorithms are applied on high-dimensional data, a critical issue is known as the curse of dimensionality. It refers to the phenomenon that data becomes sparser in high-dimensional space, adversely affecting algorithms designed for low-dimensional space~\cite{hastie2005elements}. Also, with a large number of features, learning models tend to overfit which may cause performance degradation on unseen data. Data of high dimensionality can significantly increase the memory storage requirements and computational costs for data analytics.

Dimensionality reduction is one of the most powerful tools to address the previously described issues. It can be mainly categorized into two main components: feature extraction and feature selection. Feature extraction projects the original high-dimensional features to a new feature space with low dimensionality. The newly constructed feature space is usually a linear or nonlinear combination of the original features. Feature selection, on the other hand, directly selects a subset of relevant features for model construction~\cite{guyon2003introduction,liu2007computational}.

Both feature extraction and feature selection have the advantages of improving learning performance, increasing computational efficiency, decreasing memory storage, and building better generalization models. Hence, they are both regarded as effective dimensionality reduction techniques. On one hand, for many applications where the raw input data does not contain any features understandable to a given learning algorithm, feature extraction is preferred. On the other hand, as feature extraction creates a set of new features, further analysis is problematic as we cannot retain the physical meanings of these features. In contrast, by keeping some of the original features, feature selection maintains physical meanings of the original features and gives models better readability and interpretability. Therefore, feature selection is often preferred in many applications such as text mining and genetic analysis. It should be noted that in some cases even though feature dimensionality is often not that high, feature extraction/selection still plays an essential role such as improving learning performance, preventing overfitting, and reducing computational costs.

\begin{figure}[!htbp]
\centering
\begin{minipage}{0.3\textwidth}
\centering
\subfigure[relevant feature $f_{1}$\label{fig:featureIllustration-a}]
{\includegraphics[width=\textwidth]{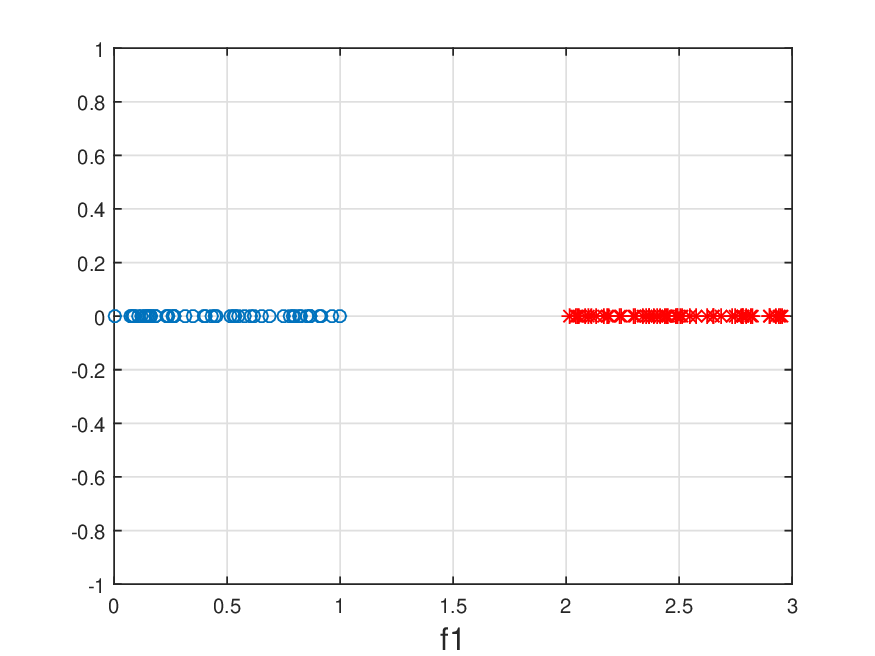}}
\end{minipage}
\begin{minipage}{0.3\textwidth}
\centering
\subfigure[redundant feature $f_{2}$\label{fig:featureIllustration-b}]
{\includegraphics[width=\textwidth]{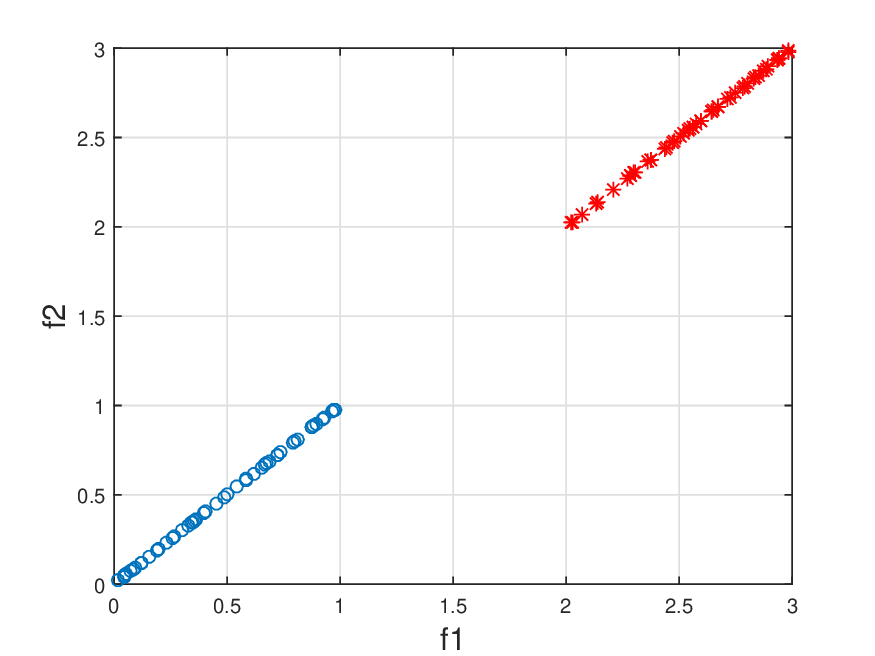}}
\end{minipage}
\begin{minipage}{0.3\textwidth}
\centering
\subfigure[irrelevant feature $f_{3}$\label{fig:featureIllustration-c}]
{\includegraphics[width=\textwidth]{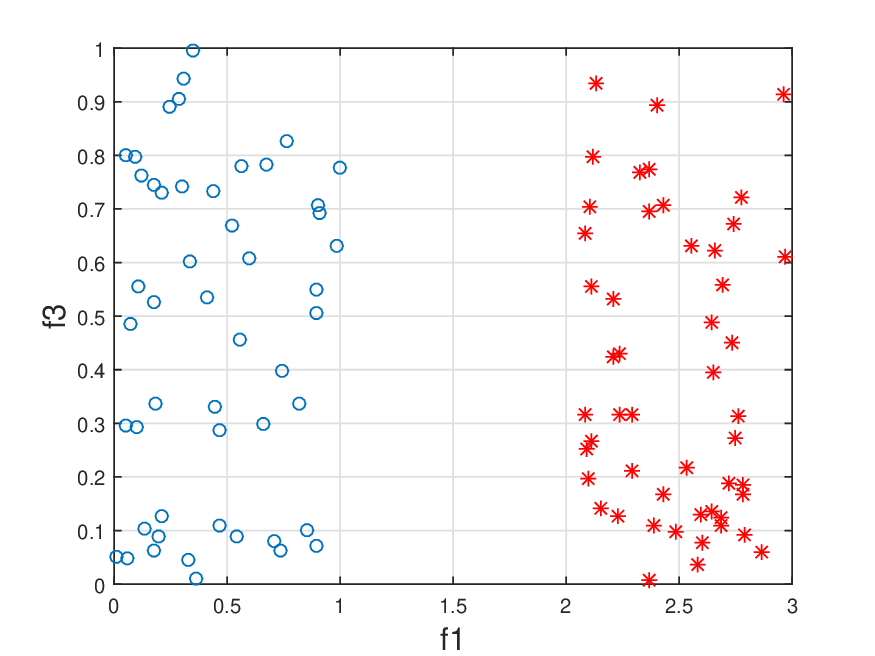}}
\end{minipage}
\centering
\caption{An illustrative example of relevant, redundant and irrelevant features.}
\vspace{-0.15in}
\label{fig:featureIllustration}
\end{figure}

Real-world data contains a lot of irrelevant, redundant and noisy features. Removing these features by feature selection reduces storage and computational cost while avoiding significant loss of information or degradation of learning performance. For example, in Fig.~\ref{fig:featureIllustration-a}, feature $f_{1}$ is a relevant feature that is able to discriminate two classes (clusters). However, given feature $f_{1}$, feature $f_{2}$ in Fig.~\ref{fig:featureIllustration-b} is redundant as $f_{2}$ is strongly correlated with $f_{1}$. In Fig.~\ref{fig:featureIllustration-c}, feature $f_{3}$ is an irrelevant feature as it cannot separate two classes (clusters) at all. Therefore, the removal of $f_{2}$ and $f_{3}$ will not negatively impact the learning performance.

\subsection{Traditional Categorization of Feature Selection Algorithms}
\subsubsection{Supervision Perspective}
According to the availability of supervision (such as class labels in classification problems), feature selection can be broadly classified as supervised, unsupervised and semi-supervised methods.

Supervised feature selection is generally designed for classification or regression problems. It aims to select a subset of features that are able to discriminate samples from different classes (classification) or to approximate the regression targets (regression). With supervision information, feature relevance is usually assessed via its correlation with the class labels or the regression target. The training phase highly depends on the selected features: after splitting the data into training and testing sets, classifiers or regression models are trained based on a subset of features selected by supervised feature selection. Note that the feature selection phase can be independent of the learning algorithms (filter methods); or it may iteratively take advantage of the learning performance of a classifier or a regression model to assess the quality of selected features so far (wrapper methods); or make use of the intrinsic structure of a learning algorithm to embed feature selection into the underlying model (embedded methods). Finally, the trained classifier or regression model predicts class labels or regression targets of unseen samples in the test set with the selected features. In the following context, for supervised methods, we mainly focus on classification problems, and use label information, supervision information interchangeably.

Unsupervised feature selection is generally designed for clustering problems. As acquiring labeled data is particularly expensive in both time and efforts, unsupervised feature selection has gained considerable attention recently. Without label information to evaluate the importance of features, unsupervised feature selection methods seek alternative criteria to define feature relevance. Different from supervised feature selection, unsupervised feature selection usually uses all instances that are available in the feature selection phase. The feature selection phase can be independent of the unsupervised learning algorithms (filter methods); or it relies on the learning algorithms to iteratively improve the quality of selected features (wrapper methods); or embed the feature selection phase into unsupervised learning algorithms (embedded methods). After the feature selection phase, it outputs the cluster structure of all data samples on the selected features by using a standard clustering algorithm~\cite{guyon2003introduction,liu2007computational,tang2014feature}.

Supervised feature selection works when sufficient label information is available while unsupervised feature selection algorithms do not require any class labels. However, in many real-world applications, we usually have a limited number of labeled data. Therefore, it is desirable to develop semi-supervised methods by exploiting both labeled and unlabeled data samples.

\vspace{-0.1in}
\subsubsection{Selection Strategy Perspective}
Concerning different selection strategies, feature selection methods can be broadly categorized as wrapper, filter and embedded methods.

Wrapper methods rely on the predictive performance of a predefined learning algorithm to evaluate the quality of selected features. Given a specific learning algorithm, a typical wrapper method performs two steps: (1) search for a subset of features; and (2) evaluate the selected features. It repeats (1) and (2) until some stopping criteria are satisfied. Feature set search component first generates a subset of features; then the learning algorithm acts as a black box to evaluate the quality of these features based on the learning performance. For example, the whole process works iteratively until such as the highest learning performance is achieved or the desired number of selected features is obtained. Then the feature subset that gives the highest learning performance is returned as the selected features. Unfortunately, a known issue of wrapper methods is that the search space for $d$ features is $2^d$, which is impractical when $d$ is very large. Therefore, different search strategies such as sequential search~\cite{guyon2003introduction}, hill-climbing search, best-first search~\cite{kohavi1997wrappers,arai2016unsupervised}, branch-and-bound search~\cite{narendra1977branch} and genetic algorithms~\cite{golberg1989genetic} are proposed to yield a local optimum learning performance. However, the search space is still extremely huge for high-dimensional datasets. As a result, wrapper methods are seldom used in practice.

Filter methods are independent of any learning algorithms. They rely on characteristics of data to assess feature importance. Filter methods are typically more computationally efficient than wrapper methods. However, due to the lack of a specific learning algorithm guiding the feature selection phase, the selected features may not be optimal for the target learning algorithms. A typical filter method consists of two steps. In the first step, feature importance is ranked according to some feature evaluation criteria. The feature importance evaluation process can be either univariate or multivariate. In the univariate scheme, each feature is ranked individually regardless of other features, while the multivariate scheme ranks multiple features in a batch way. In the second step of a typical filter method, lowly ranked features are filtered out. In the past decades, different evaluation criteria for filter methods have been proposed. Some representative criteria include feature discriminative ability to separate samples~\cite{kira1992practical,robnik2003theoretical,yang2011l2,du2013local,tang2014discriminant}, feature correlation~\cite{koller1995toward,guyon2003introduction}, mutual information~\cite{yu2003feature,peng2005feature,nguyen2014effective,shishkin2016efficient,gao2016variational}, feature ability to preserve data manifold structure~\cite{he2005laplacian,zhao2007spectral,gu2011generalized,jiang2011eigenvalue}, and feature ability to reconstruct the original data~\cite{masaeli2010convex,farahat2011efficient,li2017reconstruction}.

Embedded methods is a trade-off between filter and wrapper methods which embed the feature selection into model learning. Thus they inherit the merits of wrapper and filter methods -- (1) they include the interactions with the learning algorithm; and (2) they are far more efficient than the wrapper methods since they do not need to evaluate feature sets iteratively. The most widely used embedded methods are the regularization models which target to fit a learning model by minimizing the fitting errors and forcing feature coefficients to be small (or exact zero) simultaneously. Afterwards, both the regularization model and selected feature sets are returned as the final results.

It should be noted that some literature classifies feature selection methods into four categories (from the selection strategy perspective) by including the hybrid feature selection methods~\cite{saeys2007review,shen2012feature,ang2016supervised}. Hybrid methods can be regarded as a combination of multiple feature selection algorithms (e.g., wrapper, filter, and embedded). The main target is to tackle the instability and perturbation issues of many existing feature selection algorithms. For example, for small-sized high-dimensional data, a small perturbation on the training data may result in totally different feature selection results. By aggregating multiple selected feature subsets from different methods together, the results are more robust and hence the credibility of the selected features is enhanced.
\subsection{Feature Selection Algorithms from a Data Perspective}
\begin{figure}[!htbp]
  \centering
    \includegraphics[width=0.9\textwidth]{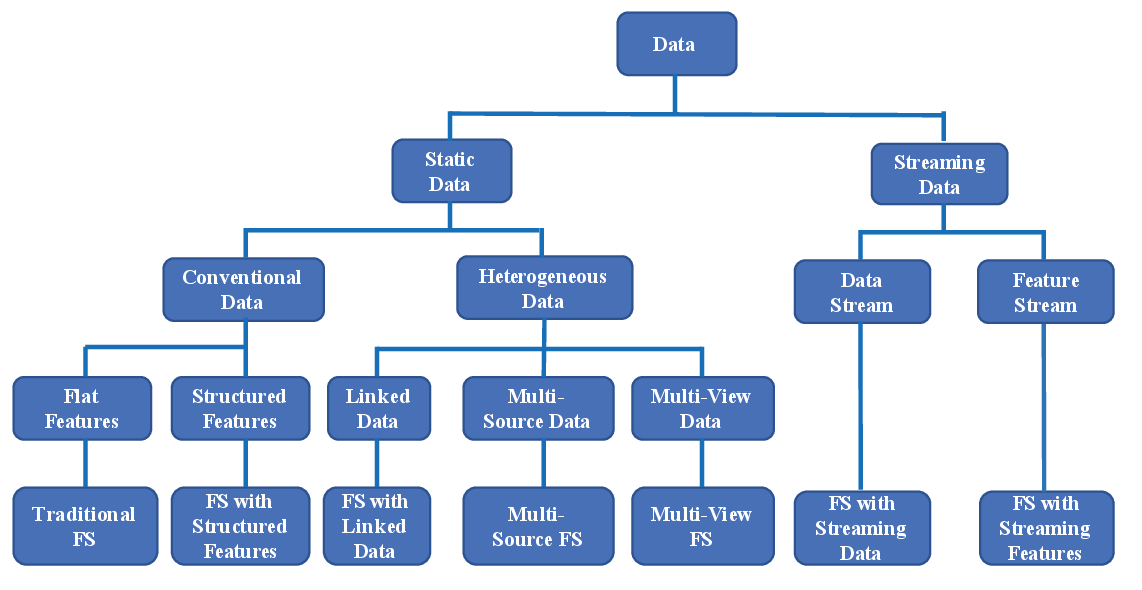}
      \caption{Feature selection algorithms from the data perspective.}
\label{fig:Categorization}
\end{figure}

The recent popularity of big data presents unique challenges for traditional feature selection~\cite{li2017challenges}, and some characteristics of big data such as velocity and variety necessitate the development of novel feature selection algorithms. Here we briefly discuss some major concerns when applying feature selection algorithms.

Streaming data and features have become more and more prevalent in real-world applications. It poses challenges to traditional feature selection algorithms, which are designed for static datasets with fixed data samples and features. For example in Twitter, new data like posts and new features like slang words are continuously being user-generated. It is impractical to apply traditional batch-mode feature selection algorithms to find relevant features from scratch when new data or new feature arrives. Moreover, the volume of data may be too large to be loaded into memory. In many cases, a single scan of data is desired as further scans is either expensive or impractical. Given the reasons mentioned above, it is appealing to apply feature selection in a streaming fashion to dynamically maintain a set of relevant features.

Most existing algorithms of feature selection are designed to handle tasks with a single data source and always assume that data is independent and identically distributed (\emph{i.i.d.}). However, data could come from multiple sources in many applications. For example, in social media, data comes from heterogeneous sources such as text, images, tags, videos. In addition, linked data is ubiquitous and presents in various forms such as user-post relations and user-user relations. The availability of multiple data sources brings unprecedented opportunities as we can leverage shared intrinsic characteristics and correlations to find more relevant features. However, challenges are also unequivocally presented. For instance, with link information, the widely adopted \emph{i.i.d.} assumption in most learning algorithms does not hold. How to appropriately utilize link information for feature selection is still a challenging problem.

Features can also exhibit certain types of structures. Some well-known structures among features are group, tree, and graph structures. When performing feature selection, if the feature structure is not taken into consideration, the intrinsic dependencies may not be captured, thus the selected features may not be suitable for the target application. Incorporating prior knowledge of feature structures can help select relevant features to improve the learning performance greatly.

The aforementioned reasons motivate the investigation of feature selection algorithms from a different view. In this survey, we revisit feature selection algorithms from a data perspective; the categorization is illustrated in Fig.~\ref{fig:Categorization}. It is shown that data consists of static data and streaming data. For the static data, it can be grouped into conventional data and heterogeneous data. In conventional data, features can either be flat or possess some inherent structures. Traditional feature selection algorithms are proposed to deal with these flat features in which features are considered to be independent. The past few decades have witnessed hundreds of feature selection algorithms. Based on their technical characteristics, we propose to classify them into four main groups, i.e., similarity based, information theoretical based, sparse learning based and statistical based methods. It should be noted that this categorization only involves filter methods and embedded methods while the wrapper methods are excluded. The reason for excluding wrapper methods is that they are computationally expensive and are usually used in specific applications. More details about these four categories will be presented later. We present other methods that cannot be fitted into these four categories, such as hybrid methods, deep learning based methods and reconstruction based methods. When features express some structures, specific feature selection algorithms are more desired. Data can be heterogeneous such that data could come from multiple sources and could be linked. Hence, we also show how new feature selection algorithms cope with these situations. Second, in the streaming settings, data arrives sequentially in a streaming fashion where the size of data instances is unknown, feature selection algorithms that make only one pass over the data is proposed accordingly. Similarly, in an orthogonal setting, features can also be generated dynamically. Streaming feature selection algorithms are designed to determine if one should accept the newly added features and remove existing but outdated features.

\subsection{Differences with Existing Surveys}
Currently, there exist some other surveys which give a summarization of feature selection algorithms, such as those in~\cite{guyon2003introduction,alelyani2013feature,chandrashekar2014survey,tang2014feature}. These studies either focus on traditional feature selection algorithms or specific learning tasks like classification and clustering. However, none of them provide a comprehensive and structured overview of traditional feature selection algorithms in conjunction with recent advances in feature selection from a data perspective. In this survey, we will introduce representative feature selection algorithms to cover all components mentioned in Fig.~\ref{fig:Categorization}. We also release a feature selection repository in Python named \emph{scikit-feature} which is built upon the widely used machine learning package \emph{scikit-learn} (http://scikit-learn.org/stable/) and two scientific computing packages \emph{Numpy} (http://www.numpy.org/) and \emph{Scipy} (http://www.scipy.org/). It includes near 40 representative feature selection algorithms. The web page of the repository is available at \url{http://featureselection.asu.edu/}.

\subsection{Organization of the Survey}
We present this survey in seven parts, and the covered topics are listed as follows:
\begin{enumerate}
  \item Traditional Feature Selection for Conventional Data (Section 2)
  \begin{enumerate}
    \item Similarity based Feature Selection Methods
    \item Information Theoretical based Feature Selection Methods
    \item Sparse Learning based Feature Selection Methods
    \item Statistical based Feature Selection Methods
    \item Other Methods
  \end{enumerate}
  \item Feature Selection with Structured Features (Section 3)
  \begin{enumerate}
    \item Feature Selection Algorithms with Group Structure Features
    \item Feature Selection Algorithms with Tree Structure Features
    \item Feature Selection Algorithms with Graph Structure Features
  \end{enumerate}
  \item Feature Selection with Heterogeneous Data (Section 4)
  \begin{enumerate}
    \item Feature Selection Algorithms with Linked Data
    \item Multi-Source Feature Selection
    \item Multi-View Feature Selection
  \end{enumerate}
  \item Feature Selection with Streaming Data (Section 5)
    \begin{enumerate}
    \item Feature Selection Algorithms with Data Streams
    \item Feature Selection Algorithms with Feature Streams
  \end{enumerate}
  \item Performance Evaluation (Section 6)
  \item Open Problems and Challenges (Section 7)
  \item Summary of the Survey (Section 8)
\end{enumerate}

\subsection{Notations}
We summarize some symbols used throughout this survey in Table~\ref{table:symbols}. We use bold uppercase characters for matrices (e.g., $\mat{A}$), bold lowercase characters for vectors (e.g., $\mat{a}$), calligraphic fonts for sets (e.g., $\mathcal{F}$). We follow the matrix settings in Matlab to represent $i$-th row of matrix $\mat{A}$ as $\mat{A}(i,:)$, $j$-th column of $\mat{A}$ as $\mat{A}(:,j)$, $(i,j)$-th entry of $\mat{A}$ as $\mat{A}(i,j)$, transpose of $\mat{A}$ as $\mat{A}'$, and trace of $\mat{A}$ as $tr(\mat{A})$. For any matrix $\mat{A}\in \mathbb{R}^{n\times d}$, its Frobenius norm is defined as $\|\mat{A}\|_{F}=\sqrt{\sum_{i=1}^{n}\sum_{j=1}^{d}\mat{A}(i,j)^{2}}$, and its $\ell_{2,1}$-norm is $\|\mat{A}\|_{2,1}=\sum_{i=1}^{n}\sqrt{\sum_{j=1}^{d}\mat{A}(i,j)^{2}}$. For any vector $\mat{a}=[a_{1},a_{2},...,a_{n}]'$, its $\ell_{2}$-norm is defined as $\|\mat{a}\|_{2}=\sqrt{\sum_{i=1}^{n}a_{i}^{2}}$, and its $\ell_{1}$-norm is $\|\mat{a}\|_{1}=\sum_{i=1}^{n}|a_{i}|$. $\mat{I}$ is an identity matrix and $\mat{1}$ is a vector whose elements are all 1's.

\begin{table}[!htbp]
\footnotesize
\centering
\begin{tabular}{|c|c|} \hline
Notations& Definitions or Descriptions \\ \hline \hline
$n$ & number of instances in the data \\ \hline
$d$ & number of features in the data\\ \hline
$k$ & number of selected features \\ \hline
$c$ & number of classes (if exist) \\ \hline
$\mathcal{F}$ & original feature set which contains $d$ features \\ \hline
$\mathcal{S}$ & selected feature set which contains $k$ selected features\\ \hline
$\{i_{1},i_{2},...,i_{k}\}$ & index of $k$ selected features in $\mathcal{S}$ \\ \hline
$f_{1},f_{2},...,f_{d}$ & $d$ original features \\ \hline
$f_{i_{1}},f_{i_{2}},...,f_{i_{k}}$ & $k$ selected features \\ \hline
$x_{1},x_{2},...,x_{n}$ & $n$ data instances \\ \hline
$\mat{f}_{1},\mat{f}_{2},...,\mat{f}_{d}$ & $d$ feature vectors corresponding to $f_{1},f_{2},...,f_{d}$ \\ \hline
$\mat{f}_{i_{1}},\mat{f}_{i_{2}},...,\mat{f}_{i_{k}}$ & $k$ feature vectors corresponding to $f_{i_{1}},f_{i_{2}},...,f_{i_{k}}$ \\ \hline
$\mat{x}_{1},\mat{x}_{2},...,\mat{x}_{n}$ & $n$ data vectors corresponding to $x_{1},x_{2},...,x_{n}$ \\ \hline
$y_{1},y_{2},...,y_{n}$ & class labels of all $n$ instances (if exist) \\ \hline
$\mat{X}\in \mathbb{R}^{n\times d}$ & data matrix with $n$ instances and $d$ features \\ \hline
$\mat{X}_{\mathcal{S}}\in \mathbb{R}^{n\times k}$ & data matrix on the selected $k$ features \\ \hline
$\mat{y}\in \mathbb{R}^{n}$ & class label vector for all $n$ instances (if exist) \\ \hline
\end{tabular}
\caption{Symbols.}
\vspace{-0.1in}
\label{table:symbols}
\end{table}

\section{Feature Selection on Conventional Data}
Over the past two decades, hundreds of feature selection algorithms have been proposed. In this section, we broadly group traditional feature selection algorithms for conventional data as similarity based, information theoretical based, sparse learning based and statistical based methods, and other methods according to the used techniques.
\vspace{-0.05in}
\subsection{Similarity based Methods}
Different feature selection algorithms exploit various types of criteria to define the relevance of features. Among them, there is a family of methods assessing feature importance by their ability to preserve data similarity. We refer them as similarity based methods. For supervised feature selection, data similarity can be derived from label information; while for unsupervised feature selection methods, most methods take advantage of different distance metric measures to obtain data similarity.

Given a dataset $\mat{X}\in \mathbb{R}^{n\times d}$ with $n$ instances and $d$ features, pairwise similarity among instances can be encoded in an affinity matrix $\mat{S}\in\mathbb{R}^{n\times n}$. Suppose that we want to select $k$ most relevant features $\mathcal{S}$, one way is to maximize their utility: $\max_{\mathcal{S}}U(\mathcal{S})$, where $U(\mathcal{S})$ denotes the utility of the feature subset $\mathcal{S}$. As algorithms in this family often evaluate features individually, the utility maximization over feature subset $\mathcal{S}$ can be further decomposed into the following form:
\begin{equation}
\small
\max_{\mathcal{S}}U(\mathcal{S})=\max_{\mathcal{S}}\sum_{f\in\mathcal{S}}U(f)=\max_{\mathcal{S}}\sum_{\mat{f}\in\mathcal{S}}\mat{\hat{f}}'\mat{\hat{S}}\mat{\hat{f}},
\label{eq:similarityFramework}
\vspace{-0.05in}
\end{equation}
where $U(f)$ is a utility function for feature $f$. $\mat{\hat{f}}$ denotes the transformation (e.g., scaling, normalization, etc) result of the original feature vector $\mat{f}$. $\mat{\hat{S}}$ is a new affinity matrix obtained from affinity matrix $\mat{S}$. The maximization problem in Eq.~(\ref{eq:similarityFramework}) shows that we would select a subset of features from $\mathcal{S}$ such that they can well preserve the data manifold structure encoded in $\mat{\hat{S}}$. This problem is usually solved by greedily selecting the top $k$ features that maximize their individual utility. Methods in this category vary in the way the affinity matrix $\mat{\hat{S}}$ is designed. We subsequently discuss some representative algorithms in this group that can be reformulated under the unified utility maximization framework.

\vspace{-0.05in}
\subsubsection{Laplacian Score}
Laplacian Score~\cite{he2005laplacian} is an unsupervised feature selection algorithm which selects features that can best preserve the data manifold structure. It consists of three phases.
First, it constructs the affinity matrix such that $\mat{S}(i,j)=e^{-\frac{\|\mat{x}_{i}-\mat{x}_{j}\|_{2}^{2}}{t}}$ if $x_{i}$ is among the $p$-nearest neighbor of $x_{j}$; otherwise $\mat{S}(i,j)=0$. Then, the diagonal matrix $\mat{D}$ is defined as $\mat{D}(i,i)=\sum_{j=1}^{n}\mat{S}(i,j)$ and the Laplacian matrix $\mat{L}$ is $\mat{L}=\mat{D}-\mat{S}$. Lastly, the Laplacian Score of each feature $f_{i}$ is computed as:
\begin{equation}
\small
laplacian\_score(f_{i})=\frac{\tilde{\mat{f}}_{i}'\mat{L}\tilde{\mat{f}}_{i}}{\tilde{\mat{f}}_{i}'\mat{D}\tilde{\mat{f}}_{i}}, \mbox{   where   } \tilde{\mat{f}_{i}}=\mat{f}_{i}-\frac{\mat{f}_{i}'\mat{D}\mat{1}}{\mat{1}'\mat{D}\mat{1}}\mat{1}.
\vspace{-0.05in}
\end{equation}
As Laplacian Score evaluates each feature individually, the task of selecting the $k$ features can be solved by greedily picking the top $k$ features with the smallest Laplacian Scores. The Laplacian Score of each feature can be reformulated as:
\begin{equation}
\small
laplacian\_score(f_{i})=1-\left(\frac{\tilde{\mat{f}}_{i}}{\|\mat{D}^{\frac{1}{2}}\tilde{\mat{f}}_{i}\|_{2}}\right)'\mat{S}\left(\frac{\tilde{\mat{f}}_{i}}{\|\mat{D}^{\frac{1}{2}}\tilde{\mat{f}}_{i}\|_{2}}\right),
\vspace{-0.05in}
\end{equation}
where $\|\mat{D}^{\frac{1}{2}}\tilde{\mat{f}}_{i}\|_{2}$ is the standard data variance of feature $f_{i}$, and the term $\tilde{\mat{f}}_{i}/\|\mat{D}^{\frac{1}{2}}\tilde{\mat{f}}_{i}\|_{2}$ is interpreted as a normalized feature vector of $\mat{f}_{i}$. Therefore, it is obvious that Laplacian Score is a special case of utility maximization in Eq.~(\ref{eq:similarityFramework}).

\vspace{-0.05in}
\subsubsection{SPEC}
SPEC~\cite{zhao2007spectral} is an extension of Laplacian Score that works for both supervised and unsupervised scenarios. For example, in the unsupervised scenario, the data similarity is measured by RBF kernel; while in the supervised scenario, data similarity can be defined by:
$
\small
\bold{S}(i,j)=\left\{
\begin{array}{ll}
\frac{1}{n_{l}}& \text{if } y_{i}=y_{j}=l\\
0& \text{otherwise}\\
\end{array}
\right.
$
, where $n_{l}$ is the number of data samples in the $l$-th class. After obtaining the affinity matrix $\mat{S}$ and the diagonal matrix $\mat{D}$, the normalized Laplacian matrix $\mat{L}_{norm}=\mat{D}^{-\frac{1}{2}}(\mat{D}-\mat{S})\mat{D}^{-\frac{1}{2}}$. The basic idea of SPEC is similar to Laplacian Score: a feature that is consistent with the data manifold structure should assign similar values to instances that are near each other. In SPEC, the feature relevance is measured by three different criteria:
\begin{equation}
\small
\begin{split}
SPEC\_score1(f_{i}) &= \hat{\mat{f}_{i}}'\gamma(\mat{L}_{norm})\hat{\mat{f}_{i}} = \sum_{j=1}^{n}\alpha_{j}^{2}\gamma(\lambda_{j})\\
SPEC\_score2(f_{i}) &= \frac{\hat{\mat{f}_{i}}'\gamma(\mat{L}_{norm})\hat{\mat{f}_{i}}}{1-(\hat{\mat{f}_{i}}'\xi_{1})^{2}} = \frac{\sum_{j=2}^{n}\alpha_{j}^{2}\gamma(\lambda_{j})}{\sum_{j=2}^{n}\alpha_{j}^{2}}\\
SPEC\_score3(f_{i}) &= \sum_{j=1}^{m}(\gamma(2)-\gamma(\lambda_{j}))\alpha_{j}^{2}.
\vspace{-0.1in}
\end{split}
\label{eq:SPECCriteria}
\end{equation}
In the above equations, $\hat{\mat{f}_{i}}=\mat{D}^{\frac{1}{2}}\mat{f}_{i}/\|\mat{D}^{\frac{1}{2}}\mat{f}_{i}\|_{2}$; $(\lambda_{j},\xi_{j})$ is the $j$-th eigenpair of the normalized Laplacian matrix $\mat{L}_{norm}$; $\alpha_{j}=\cos\theta_{j}$, $\theta_{j}$ is the angle between $\xi_{j}$ and $\mat{f}_{i}$; $\gamma(.)$ is an increasing function to penalize high frequency components of the eigensystem to reduce noise. If the data is noise free, the function $\gamma(.)$ can be removed and $\gamma(x)=x$. When the second evaluation criterion $SPEC\_score2(f_{i})$ is used, SPEC is equivalent to the Laplacian Score. For $SPEC\_score3(f_{i})$, it uses the top $m$ eigenpairs to evaluate the importance of feature $f_{i}$.

All these three criteria can be reduced to the the unified similarity based feature selection framework in Eq.~(\ref{eq:similarityFramework}) by setting $\mat{\hat{f}}_{i}$ as $\mat{f}_{i}\|\mat{D}^{\frac{1}{2}}\mat{f}_{i}\|_{2}$, $(\mat{f}_{i}-\mu\mat{1})/\|\mat{D}^{\frac{1}{2}}\mat{f}_{i}\|_{2}$, $\mat{f}_{i}\|\mat{D}^{\frac{1}{2}}\mat{f}_{i}\|_{2}$; and $\mat{\hat{S}}$ as $\mat{D}^{\frac{1}{2}}\mat{U}(\mat{I}-\gamma(\mat{\Sigma}))\mat{U}'\mat{D}^{\frac{1}{2}}$, $\mat{D}^{\frac{1}{2}}\mat{U}(\mat{I}-\gamma(\mat{\Sigma}))\mat{U}'\mat{D}^{\frac{1}{2}}$, $\mat{D}^{\frac{1}{2}}\mat{U}_{m}(\gamma(2\mat{I})-\gamma(\mat{\Sigma}_{m}))\mat{U}_{m}'\mat{D}^{\frac{1}{2}}$ in SPEC\_score1, SPEC\_score2, SPEC\_score3, respectively. $\mat{U}$ and $\mat{\Sigma}$ are the singular vectors and singular values of the normalized Laplacian matrix $\mat{L}_{norm}$.

\vspace{-0.05in}
\subsubsection{Fisher Score}
Fisher Score~\cite{duda2012pattern} is a supervised feature selection algorithm. It selects features such that the feature values of samples within the same class are similar while the feature values of samples from different classes are dissimilar. The Fisher Score of each feature $f_{i}$ is evaluated as follows:

\begin{equation}
\small
fisher\_score(f_{i})=\frac{\sum_{j=1}^{c}n_{j}(\mu_{ij}-\mu_{i})^{2}}{\sum_{j=1}^{c}n_{j}\sigma_{ij}^{2}},
\vspace{-0.05in}
\end{equation}
where $n_{j}$, $\mu_{i}$, $\mu_{ij}$ and $\sigma_{ij}^{2}$ indicate the number of samples in class $j$, mean value of feature $f_{i}$, mean value of feature $f_{i}$ for samples in class $j$, variance value of feature $f_{i}$ for samples in class $j$, respectively. Similar to Laplacian Score, the top $k$ features can be obtained by greedily selecting the features with the largest Fisher Scores.

According to~\cite{he2005laplacian}, Fisher Score can be considered as a special case of Laplacian Score as long as the affinity matrix is
$\bold{S}(i,j)=\left\{
\begin{array}{ll}
\frac{1}{n_{l}}& \text{if } y_{i}=y_{j}=l\\
0& \text{otherwise,}\\
\end{array}
\right.
$. In this way, the relationship between Fisher Score and Laplacian Score is $fisher\_score(f_{i})=1-\frac{1}{laplacian\_score(f_{i})}$. Hence, the computation of Fisher Score can also be reduced to the unified utility maximization framework.

\vspace{-0.05in}
\subsubsection{Trace Ratio Criterion}
The trace ratio criterion~\cite{nie2008trace} directly selects the global optimal feature subset based on the corresponding score, which is computed by a trace ratio norm. It builds two affinity matrices $\mat{S}_{w}$ and $\mat{S}_{b}$ to characterize within-class and between-class data similarity. Let $\mat{W}=[\mat{w}_{i_{1}},\mat{w}_{i_{2}},...,\mat{w}_{i_{k}}]\in\mathbb{R}^{d\times k}$ be the selection indicator matrix such that only the $i_{j}$-th entry in $\mat{w}_{i_{j}}$ is 1 and all the other entries are 0. With these, the trace ratio score of the selected $k$ features in $\mathcal{S}$ is:
\begin{equation}
\small
trace\_ratio(\mathcal{S})=\frac{tr(\mat{W}'\mat{X}'\mat{L}_{b}\mat{X}\mat{W})}{tr(\mat{W}'\mat{X}'\mat{L}_{w}\mat{X}\mat{W})},
\end{equation}
where $\mat{L}_{b}$ and $\mat{L}_{w}$ are Laplacian matrices of $\mat{S}_{a}$ and $\mat{S}_{b}$ respectively. The basic idea is to maximize the data similarity for instances from the same class while minimize the data similarity for instances from different classes. However, the trace ratio problem is difficult to solve as it does not have a closed-form solution. Hence, the trace ratio problem is often converted into a more tractable format called the ratio trace problem by maximizing $tr[(\mat{W}'\mat{X}'\mat{L}_{w}\mat{X}\mat{W})^{-1}(\mat{W}'\mat{X}'\mat{L}_{b}\mat{X}\mat{W})]$. As an alternative,~\cite{wang2007trace} propose an iterative algorithm called ITR to solve the trace ratio problem directly and was later applied in trace ratio feature selection~\cite{nie2008trace}.

Different $\mat{S}_{b}$ and $\mat{S}_{w}$ lead to different feature selection algorithms such as batch-mode Lalpacian Score and batch-mode Fisher Score. For example, in batch-mode Fisher Score, the within-class data similarity and the between-class data similarity are $
\mat{S}_w(i,j)=\left\{
\begin{array}{ll}
1/n_{l}& \text{if } y_{i}=y_{j}=l\\
0& \text{otherwise}\\
\end{array}
\right.
$
and
$
\mat{S}_b(i,j)=\left\{
\begin{array}{ll}
1/n-1/n_{l}& \text{if } y_{i}=y_{j}=l\\
1/n& \text{otherwise}\\
\end{array}
\right.
$ respectively.
Therefore, maximizing the trace ratio criterion is equivalent to maximizing
$
\frac{\sum_{s=1}^{k}\mat{f}_{i_{s}}'\mat{S}_{w}\mat{f}_{i_{s}}}{\sum_{s=1}^{k}\mat{f}_{i_{s}}'\mat{f}_{i_{s}}}=\frac{\mat{X}_{\mathcal{S}}'\mat{S}_{w}\mat{X}_{\mathcal{S}}}{\mat{X}_{\mathcal{S}}'\mat{X}_{\mathcal{S}}}.
$
Since $\mat{X}_{\mathcal{S}}'\mat{X}_{\mathcal{S}}$ is constant, it can be further reduced to the unified similarity based feature selection framework by setting $\mat{\hat{f}}=\mat{f}/\|\mat{f}\|_{2}$ and $\mat{\hat{S}}=\mat{S}_{w}$. On the other hand in batch-mode Laplacian Score, the within-class data similarity and the between-class data similarity are $
\mat{S}_w(i,j)=\left\{
\begin{array}{ll}
e^{-\frac{\|\mat{x}_{i}-\mat{x}_{j}\|_{2}^{2}}{t}}& \text{if } \mat{x}_{i}\in \mathcal{N}_{p}(\mat{x}_{j}) \text{ or }\mat{x}_{j}\in \mathcal{N}_{p}(\mat{x}_{i})\\
0& \text{otherwise}\\
\end{array}
\right.
$
and
$
\mat{S}_{b}=(\mat{1}'\mat{D}_{w}\mat{1})^{-1}\mat{D}_{w}\mat{1}\mat{1}'\mat{D}_{w}
$ respectively. In this case, the trace ratio criterion score is
$
\frac{tr(\mat{W}'\mat{X}'\mat{L}_{b}\mat{X}\mat{W})}{tr(\mat{W}'\mat{X}'\mat{L}_{w}\mat{X}\mat{W})}=
\frac{\sum_{s=1}^{k}\mat{f}_{i_{s}}'\mat{D}_{w}\mat{f}_{i_{s}}}{\sum_{s=1}^{k}\mat{f}_{i_{s}}'(\mat{D}_{w}-\mat{S}_{w})\mat{f}_{i_{s}}}
$. Therefore, maximizing the trace ratio criterion is also equivalent to solving the unified maximization problem in Eq.~(\ref{eq:similarityFramework}) where $\mat{\hat{f}}=\mat{f}/\|\mat{D}_{w}^{\frac{1}{2}}\mat{f}\|_{2}$ and $\mat{\hat{S}}=\mat{S}_{w}$.

\subsubsection{ReliefF}
ReliefF~\cite{robnik2003theoretical} selects features to separate instances from different classes. Assume that $l$ data instances are randomly selected among all $n$ instances, then the feature score of $f_{i}$ in ReliefF is defined as follows:
\begin{equation}
\small
\begin{split}
ReliefF\_score(f_{i})=&\frac{1}{c}\sum_{j=1}^{l}(-\frac{1}{m_{j}}\sum_{x_{r}\in \text{\emph{NH}}(j)}d(\mat{X}(j,i)-\mat{X}(r,i))\\
+&\sum_{y\neq y_{j}}\frac{1}{h_{jy}}\frac{p(y)}{1-p(y)}\sum_{x_{r}\in \text{\emph{NM}}(j,y)}d(\mat{X}(j,i)-\mat{X}(r,i))),
\end{split}
\vspace{-0.1in}
\end{equation}
where \emph{NH}$(j)$ and \emph{NM}$(j,y)$ are the nearest instances of $x_{j}$ in the same class and in class $y$, respectively. Their sizes are $m_{j}$ and $h_{jy}$, respectively. $p(y)$ is the ratio of instances in class $y$.

ReliefF is equivalent to selecting features that preserve a special form of data similarity matrix. Assume that the dataset has the same number of instances in each of the $c$ classes and there are $q$ instances in both $NM(j)$ and $NH(j,y)$. Then according to~\cite{zhao2007spectral}, the ReliefF feature selection can be reduced to the utility maximization framework in Eq.~(\ref{eq:similarityFramework}).

\vspace{-0.02in}
\paragraph{\textbf{Discussion}: Similarity based feature selection algorithms have demonstrated with excellent performance in both supervised and unsupervised learning problems. This category of methods is straightforward and simple as the computation focuses on building an affinity matrix, and afterwards, the scores of features can be obtained. Also, these methods are independent of any learning algorithms and the selected features are suitable for many subsequent learning tasks. However, one drawback of these methods is that most of them cannot handle feature redundancy. In other words, they may repeatedly find highly correlated features during the selection phase}
\vspace{-0.05in}
\subsection{Information Theoretical based Methods}
A large family of existing feature selection algorithms is information theoretical based methods. Algorithms in this family exploit different heuristic filter criteria to measure the importance of features. As indicated in~\cite{duda2012pattern}, many hand-designed information theoretic criteria are proposed to maximize feature relevance and minimize feature redundancy. Since the relevance of a feature is usually measured by its correlation with class labels, most algorithms in this family are performed in a supervised way. In addition, most information theoretic concepts can only be applied to discrete variables. Therefore, feature selection algorithms in this family can only work with discrete data. For continuous feature values, some data discretization techniques are required beforehand. Two decades of research on information theoretic criteria can be unified in a conditional likelihood maximization framework~\cite{brown2012conditional}. In this subsection, we introduce some representative algorithms in this family. We first give a brief introduction about basic information theoretic concepts.

The concept of \emph{entropy} measures the uncertainty of a discrete random variable. The entropy of a discrete random variable $X$ is defined as follows:
\begin{equation}
\small
H(X)=-\sum_{x_{i}\in X}P(x_{i})log(P(x_{i})),
\vspace{-0.1in}
\end{equation}
where $x_{i}$ denotes a specific value of random variable $X$, $P(x_{i})$ denotes the probability of $x_{i}$ over all possible values of $X$.

Second, the \emph{conditional entropy} of $X$ given another discrete random variable $Y$ is:
\begin{equation}
\small
H(X|Y)=-\sum_{y_{j}\in Y}P(y_{j})\sum_{x_{i} \in X}P(x_{i}|y_{j})log(P(x_{i}|y_{j})),
\vspace{-0.1in}
\end{equation}
where $P(y_{i})$ is the prior probability of $y_{i}$, while $P(x_{i}|y_{j})$ is the conditional probability of $x_{i}$ given $y_{j}$. It shows the uncertainty of $X$ given $Y$.

Then, \emph{information gain} or \emph{mutual information} between $X$ and $Y$ is used to measure the amount of information shared by $X$ and $Y$ together:
\begin{equation}
\small
I(X;Y)=H(X)-H(X|Y)=\sum_{x_{i}\in X}\sum_{y_{j}\in Y}P(x_{i},y_{j})log\frac{P(x_{i},y_{j})}{P(x_{i})P(y_{j})},
\vspace{-0.05in}
\end{equation}
where $P(x_{i},y_{j})$ is the joint probability of $x_{i}$ and $y_{j}$. Information gain is symmetric such that $I(X;Y)=I(Y;X)$, and is zero if the discrete variables $X$ and $Y$ are independent.

At last, \emph{conditional information gain} (or \emph{conditional mutual information}) of discrete variables $X$ and $Y$ given a third discrete variable $Z$ is given as follows:
\begin{equation}
\small
I(X;Y|Z)=H(X|Z)-H(X|Y,Z)=\sum_{z_{k}\in Z}P(z_{k})\sum_{x_{i}\in X}\sum_{y_{j}\in Y}P(x_{i},y_{j}|z_{k})log\frac{P(x_{i},y_{j}|z_{k})}{P(x_{i}|z_{k})P(y_{j}|z_{k})}.
\vspace{-0.05in}
\end{equation}
It shows the amount of mutual information shared by $X$ and $Y$ given $Z$.

Searching for the global best set of features is NP-hard, thus most algorithms exploit heuristic sequential search approaches to add/remove features one by one. In this survey, we explain the feature selection problem by forward sequential search such that features are added into the selected feature set one by one. We denote $\mathcal{S}$ as the current selected feature set that is initially empty. $Y$ represents the class labels. $X_{j}\in \mathcal{S}$ is a specific feature in the current $\mathcal{S}$. $J(.)$ is a feature selection criterion (score) where, generally, the higher the value of $J(X_{k})$, the more important the feature $X_{k}$ is. In the unified conditional likelihood maximization feature selection framework, the selection criterion (score) for a new unselected feature $X_{k}$ is given as follows:
\begin{equation}
\small
J_{\text{CMI}}(X_{k})=I(X_{k};Y)+\sum_{X_{j}\in\mathcal{S}}g[I(X_{j};X_{k}),I(X_{j};X_{k}|Y)],
\label{eq:cmi}
\vspace{-0.05in}
\end{equation}
where $g(.)$ is a function w.r.t. two variables $I(X_{j};X_{k})$ and $I(X_{j};X_{k}|Y)$. If $g(.)$ is a linear function w.r.t. these two variables, it is referred as a criterion by linear combinations of Shannon information terms such that:
\begin{equation}
\small
J_{\text{CMI}}(X_{k})=I(X_{k};Y)-\beta\sum_{X_{j}\in\mathcal{S}}I(X_{j};X_{k})+\lambda\sum_{X_{j}\in\mathcal{S}}I(X_{j};X_{k}|Y).
\label{eq:CMI}
\vspace{-0.05in}
\end{equation}
where $\beta$ and $\lambda$ are two nonnegative parameters between zero and one. On the other hand, if $g(.)$ is a non-linear function w.r.t. these two variables, it is referred as a criterion by non-linear combination of Shannon information terms.

\subsubsection{Mutual Information Maximization (Information Gain)}
Mutual Information Maximization (MIM) (a.k.a. Information Gain)~\cite{lewis1992feature} measures the importance of a feature by its correlation with class labels. It assumes that when a feature has a strong correlation with the class label, it can help achieve good classification performance. The Mutual Information score for feature $X_{k}$ is:
\begin{equation}
\small
J_{\text{MIM}}(X_{k})=I(X_{k};Y).
\vspace{-0.05in}
\end{equation}
It can be observed that in MIM, the scores of features are assessed individually. Therefore, only the feature correlation is considered while the feature redundancy is completely ignored. After it obtains the MIM feature scores for all features, we choose the features with the highest feature scores and add them to the selected feature set. The process repeats until the desired number of selected features is obtained.

It can also be observed that MIM is a special case of linear combination of Shannon information terms in Eq.~(\ref{eq:CMI}) where both $\beta$ and $\lambda$ are equal to zero.

\subsubsection{Mutual Information Feature Selection}
A limitation of MIM criterion is that it assumes that features are independent of each other. In reality, good features should not only be strongly correlated with class labels but also should not be highly correlated with each other. In other words, the correlation between features should be minimized. Mutual Information Feature Selection (MIFS)~\cite{battiti1994using} considers both the feature relevance and feature redundancy in the feature selection phase, the feature score for a new unselected feature $X_{k}$ can be formulated as follows:
\begin{equation}
\small
J_{\text{MIFS}}(X_{k})=I(X_{k};Y)-\beta\sum_{X_{j}\in\mathcal{S}}I(X_{k};X_{j}).
\vspace{-0.05in}
\end{equation}
In MIFS, the feature relevance is evaluated by $I(X_{k};Y)$, while the second term penalizes features that have a high mutual information with the currently selected features such that feature redundancy is minimized.

MIFS can also be reduced to be a special case of the linear combination of Shannon information terms in Eq.~(\ref{eq:CMI}) where $\beta$ is between zero and one, and $\lambda$ is zero.

\subsubsection{Minimum Redundancy Maximum Relevance}
~\cite{peng2005feature} proposes a Minimum Redundancy Maximum Relevance (MRMR) criterion to set the value of $\beta$ to be the reverse of the number of selected features:
\begin{equation}
\small
J_{\text{MRMR}}(X_{k})=I(X_{k};Y)-\frac{1}{|\mathcal{S}|}\sum_{X_{j}\in\mathcal{S}}I(X_{k};X_{j}).
\vspace{-0.05in}
\end{equation}
Hence, with more selected features, the effect of feature redundancy is gradually reduced. The intuition is that with more non-redundant features selected, it becomes more difficult for new features to be redundant to the features that have already been in $\mathcal{S}$. In~\cite{brown2012conditional}, it gives another interpretation that the pairwise independence between features becomes stronger as more features are added to $\mathcal{S}$, possibly because of noise information in the data.

MRMR is also strongly linked to the Conditional likelihood maximization framework if we iteratively revise the value of $\beta$ to be $\frac{1}{|\mathcal{S}|}$, and set the other parameter $\lambda$ to be zero.

\vspace{-0.05in}
\subsubsection{Conditional Infomax Feature Extraction}
Some studies~\cite{lin2006conditional,el2008powerful,guo2009gait} show that in contrast to minimize the feature redundancy, the conditional redundancy between unselected features and already selected features given class labels should also be maximized. In other words, as long as the feature redundancy given class labels is stronger than the intra-feature redundancy, the feature selection will be affected negatively. A typical feature selection under this argument is Conditional Infomax Feature Extraction (CIFE)~\cite{lin2006conditional}, in which the feature score for a new unselected feature $X_{k}$ is:
\begin{equation}
\small
J_{\text{CIFE}}(X_{k})=I(X_{k};Y)-\sum_{X_{j}\in\mathcal{S}}I(X_{j};X_{k})+\sum_{X_{j}\in\mathcal{S}}I(X_{j};X_{k}|Y).
\end{equation}
Compared with MIFS, it adds a third term $\sum_{X_{j}\in\mathcal{S}}I(X_{j};X_{k}|Y)$ to maximize the conditional redundancy. Also, CIFE is a special case of the linear combination of Shannon information terms by setting both $\beta$ and $\gamma$ to be 1.

\subsubsection{Joint Mutual Information}
MIFS and MRMR reduce feature redundancy in the feature selection process. An alternative criterion, Joint Mutual Information~\cite{yang1999data,meyer2008information} is proposed to increase the complementary information that is shared between unselected features and selected features given the class labels. The feature selection criterion is listed as follows:
\begin{equation}
\small
J_{\text{JMI}}(X_{k})=\sum_{X_{j}\in \mathcal{S}}I(X_{k},X_{j};Y).
\end{equation}
The basic idea of JMI is that we should include new features that are complementary to the existing features given the class labels.

JMI cannot be directly reduced to the condition likelihood maximization framework. In~\cite{brown2012conditional}, the authors demonstrate that with simple manipulations, the JMI criterion can be re-written as:
\begin{equation}
\small
J_{\text{JMI}}(X_{k})=I(X_{k};Y)-\frac{1}{|\mathcal{S}|}\sum_{X_{j}\in\mathcal{S}}I(X_{j};X_{k})+\frac{1}{|\mathcal{S}|}\sum_{X_{j}\in\mathcal{S}}I(X_{j};X_{k}|Y).
\end{equation}
Therefore, it is also a special case of the linear combination of Shannon information terms by iteratively setting $\beta$ and $\lambda$ to be $\frac{1}{|\mathcal{S}|}$.

\subsubsection{Conditional Mutual Information Maximization}
Previously mentioned criteria could be reduced to a linear combination of Shannon information terms. Next, we show some other algorithms that can only be reduced to a non-linear combination of Shannon information terms. Among them, Conditional Mutual Information Maximization (CMIM)~\cite{vidal2003object,fleuret2004fast} iteratively selects features which maximize the mutual information with the class labels given the selected features so far. Mathematically, during the selection phase, the feature score for each new unselected feature $X_{k}$ can be formulated as follows:
\begin{equation}
\small
J_{\text{CMIM}}(X_{k})=\min_{X_{j}\in \mathcal{S}}[I(X_{k};Y|X_{j})].
\end{equation}
Note that the value of $I(X_{k};Y|X_{j})$ is small if $X_{k}$ is not strongly correlated with the class label $Y$ or if $X_{k}$ is redundant when $\mathcal{S}$ is known. By selecting the feature that maximizes this minimum value, it can guarantee that the selected feature has a strong predictive ability, and it can reduce the redundancy w.r.t. the selected features.

The CMIM criterion is equivalent to the following form after some derivations:
\begin{equation}
\small
J_{\text{CMIM}}(X_{k})=I(X_{k};Y)-\max_{X_{j}\in\mathcal{S}}[I(X_{j};X_{k})-I(X_{j};X_{k}|Y)].
\label{eq:CMIM}
\end{equation}
Therefore, CMIM is also a special case of the conditional likelihood maximization framework in Eq.~(\ref{eq:cmi}).

\subsubsection{Informative Fragments}
In~\cite{vidal2003object}, the authors propose a feature selection criterion called Informative Fragments (IF). The feature score of each new unselected features is given as:
\begin{equation}
\small
J_{\text{IF}}(X_{k})=\min_{X_{j}\in\mathcal{S}}[I(X_{j}X_{k};Y)-I(X_{j};Y)].
\end{equation}
The intuition behind Informative Fragments is that the addition of the new feature $X_{k}$ should maximize the value of conditional mutual information between $X_{k}$ and existing features in $\mathcal{S}$ over the mutual information between $X_{j}$ and $Y$. An interesting phenomenon of IF is that with the chain rule that $I(X_{k}X_{j};Y)=I(X_{j};Y)+I(X_{k};Y|X_{j})$, IF has the equivalent form as CMIM. Hence, it can also be reduced to the general framework in Eq.~(\ref{eq:cmi}).

\subsubsection{Interaction Capping}
Interaction Capping~\cite{jakulin2005machine} is a similar feature selection criterion as CMIM in Eq.~(\ref{eq:CMIM}), it restricts the term $I(X_{j};X_{k})-I(X_{j};X_{k}|Y)$ to be nonnegative:
\begin{equation}
\small
J_{\text{CMIM}}(X_{k})=I(X_{k};Y)-\sum_{X_{j}\in\mathcal{S}}\max[0,I(X_{j};X_{k})-I(X_{j};X_{k}|Y)].
\end{equation}
Apparently, it is a special case of non-linear combination of Shannon information terms by setting the function $g(.)$ to be $-\max[0,I(X_{j};X_{k})-I(X_{j};X_{k}|Y)]$.

\subsubsection{Double Input Symmetrical Relevance}
Another class of information theoretical based methods such as Double Input Symmetrical Relevance (DISR)~\cite{meyer2006use} exploits normalization techniques to normalize mutual information~\cite{guyon2008feature}:
\begin{equation}
\small
J_{\text{DISR}}(X_{k})=\sum_{X_{j}\in \mathcal{S}}\frac{I(X_{j}X_{k};Y)}{H(X_{j}X_{k}Y)}.
\end{equation}
It is easy to validate that DISR is a non-linear combination of Shannon information terms and can be reduced to the conditional likelihood maximization framework.

\subsubsection{Fast Correlation Based Filter}

There are other information theoretical based feature selection methods that cannot be simply reduced to the unified conditional likelihood maximization framework. Fast Correlation Based Filter (FCBF)~\cite{yu2003feature} is an example that exploits feature-class correlation and feature-feature correlation simultaneously. The algorithm works as follows: (1) given a predefined threshold $\delta$, it selects a subset of features $\mathcal{S}$ that are highly correlated with the class labels with $SU\geq \delta$, where $SU$ is the symmetric uncertainty. The $SU$ between a set of features $X_{\mathcal{S}}$ and the class label $Y$ is given as follows:
\begin{equation}
\small
SU(X_{\mathcal{S}},Y)=2\frac{I(X_{\mathcal{S}};Y)}{H(X_{\mathcal{S}})+H(Y)}.
\end{equation}
A specific feature $X_{k}$ is called predominant iff $SU(X_{k},Y)\geq \delta$ and there does not exist a feature $X_{j}\in\mathcal{S}$ $(j\neq k)$ such that $SU(X_{j},X_{k})\geq SU(X_{k},Y)$. Feature $X_{j}$ is considered to be redundant to feature $X_{k}$ if $SU(X_{j},X_{k})\geq SU(X_{k},Y)$;  (2) the set of redundant features is denoted as $\mathcal{S}_{P_{i}}$, which will be further split into $\mathcal{S}_{P_{i}}^{+}$ and $\mathcal{S}_{P_{i}}^{-}$ where they contain redundant features to feature $X_{k}$ with $SU(X_{j},Y)>SU(X_{k},Y)$ and $SU(X_{j},Y)<SU(X_{k},Y)$, respectively; and (3) different heuristics are applied on $\mathcal{S}_{P}$, $\mathcal{S}_{P_{i}}^{+}$ and $\mathcal{S}_{P_{i}}^{-}$ to remove redundant features and keep the features that are most relevant to the class labels.

\vspace{-0.05in}
\paragraph{\textbf{Discussion}: Unlike similarity based feature selection algorithms that fail to tackle feature redundancy, most aforementioned information theoretical based feature selection algorithms can be unified in a probabilistic framework that considers both ``feature relevance" and ``feature redundancy". Meanwhile, similar as similarity based methods, this category of methods is independent of any learning algorithms and hence are generalizable. However, most of the existing information theoretical based feature selection methods can only work in a supervised scenario. Without the guide of class labels, it is still not clear how to assess the importance of features. In addition, these methods can only handle discrete data and continuous numerical variables require discretization preprocessing beforehand}

\vspace{-0.05in}
\subsection{Sparse Learning based Methods}
The third type of methods is sparse learning based methods which aim to minimize the fitting errors along with some sparse regularization terms. The sparse regularizer forces many feature coefficients to be small, or exactly zero, and then the corresponding features can be simply eliminated. Sparse learning based methods have received considerable attention in recent years due to their good performance and interpretability. In the following parts, we review some representative sparse learning based feature selection methods from both supervised and unsupervised perspectives.
\vspace{-0.05in}
\subsubsection{Feature Selection with $\ell_{p}$-norm Regularizer}

First, we consider the binary classification or univariate regression problem. To achieve feature selection, the $\ell_{p}$-norm sparsity-induced penalty term is added on the classification or regression model, where $0\leq p \leq 1$. Let $\mat{w}$ denotes the feature coefficient, then the objective function for feature selection is:
\begin{equation}
\small
\min_{\mat{w}}\, loss(\mat{w};\mat{X},\mat{y})+\alpha \, \|\mat{w}\|_{p},
\label{eq:fs_regularization}
\vspace{-0.05in}
\end{equation}
where $loss(.)$ is a loss function, and some widely used loss functions $loss(.)$ include least squares loss, hinge loss and logistic loss. $\|\mat{w}\|_{p}=(\sum_{i=1}^{d}\|w_{i}\|^{p})^{\frac{1}{p}}$ is a sparse regularization term, and $\alpha$ is a regularization parameter to balance the contribution of the loss function and the sparse regularization term for feature selection.

Typically when $p=0$, the $\ell_{0}$-norm regularization term directly seeks for the optimal set of nonzero entries (features) for the model learning. However, the optimization problem is naturally an integer programming problem and is difficult to solve. Therefore, it is often relaxed to a $\ell_{1}$-norm regularization problem, which is regarded as the tightest convex relaxation of the $\ell_{0}$-norm. One main advantage of $\ell_{1}$-norm regularization (LASSO)~\cite{tibshirani1996regression} is that it forces many feature coefficients to become smaller and, in some cases, exactly zero. This property makes it suitable for feature selection, as we can select features whose corresponding feature weights are large, which motivates a surge of $\ell_{1}$-norm regularized feature selection methods~\cite{zhu20041,xu2014gradient,wei2016nonlinear,wei2016unsupervised,hara2017enumerate}. Also, the sparse vector $\mat{w}$ enables the ranking of features. Normally, the higher the value, the more important the corresponding feature is.
\vspace{-0.05in}
\subsubsection{Feature Selection with $\ell_{p,q}$-norm Regularizer}

Here, we discuss how to perform feature selection for the general multi-class classification or multivariate regression problems. The problem is more difficult because of the multiple classes and multivariate regression targets, and we would like the feature selection phase to be consistent over multiple targets. In other words, we want multiple predictive models for different targets to share the same parameter sparsity patterns -- each feature either has small scores or large scores for all targets. This problem can be generally solved by the $\ell_{p,q}$-norm sparsity-induced regularization term, where $p>1$ (most existing work focus on $p=2$ or $\infty$) and $0\leq q \leq 1$ (most existing work focus on $q=1$ or $0$). Assume that $\mat{X}$ denotes the data matrix, and $\mat{Y}$ denotes the one-hot label indicator matrix. Then the model is formulated as follows:
\begin{equation}
\small
\min_{\mat{W}}\, loss(\mat{W};\mat{X},\mat{y})+\alpha\|\mat{W}\|_{p,q},
\vspace{-0.05in}
\end{equation}
where $\|\mat{W}\|_{p,q}=(\sum_{j=1}^{c}(\sum_{i=1}^{d}|\mat{W}(i,j)|^{p})^{\frac{q}{p}})^{\frac{1}{q}}$; and the parameter $\alpha$ is used to control the contribution of the loss function and the sparsity-induced regularization term. Then the features can be ranked according to the value of $\|\mat{W}(i,:)\|_{2}^{2}({i=1,...,d})$, the higher the value, the more important the feature is.
\vspace{-0.05in}
\paragraph{Case 1: $p=2$, $q=0$} To find relevant features across multiple targets, an intuitive way is to use discrete optimization through the $\ell_{2,0}$-norm regularization. The optimization problem with the $\ell_{2,0}$-norm regularization term can be reformulated as follows:
\begin{equation}
\small
\min_{\mat{W}}\, loss(\mat{W};\mat{X},\mat{y})\, \, s.t. \|\mat{W}\|_{2,0}\leq k.
\vspace{-0.05in}
\end{equation}
However, solving the above optimization problem has been proven to be NP-hard, and also, due to its discrete nature, the objective function is also not convex. To solve it, a variation of Alternating Direction Method could be leveraged to seek for a local optimal solution~\cite{cai2013exact,gu2012locality}. In~\cite{zhang2014feature}, the authors provide two algorithms, proximal gradient algorithm and rank-one update algorithm to solve this discrete selection problem.
\vspace{-0.05in}
\paragraph{Case 2: $p=2$, $0<q<1$} The above sparsity-reduced regularization term is inherently discrete and hard to solve. In~\cite{peng2016direct,peng2017general}, the authors propose a more general framework to directly optimize the sparsity-reduced regularization when $0<q<1$ and provided efficient iterative algorithm with guaranteed convergence rate.
\vspace{-0.05in}
\paragraph{Case 3: $p=2$, $q=1$} Although the $\ell_{2,0}$-norm is more desired for feature sparsity, however, it is inherently non-convex and non-smooth. Hence, the $\ell_{2,1}$-norm regularization is preferred and widely used in different scenarios such as multi-task learning~\cite{obozinski2007joint,zhang2008flexible}, anomaly detection~\cite{li2017residual,wu2017gleaning} and crowdsourcing~\cite{zhou2017learning}. Many $\ell_{2,1}$-norm regularization based feature selection methods have been proposed over the past decade~\cite{zhao2010efficient,gu2011joint,yang2011l2,hou2011feature,li2012unsupervised,qian2013robust,shi2014robust,liu2014global,du2015unsupervised,jian2016multi,liu2016consensus,nie2016unsupervised,zhu2016coupled,li2017toward}. Similar to $\ell_{1}$-norm regularization, $\ell_{2,1}$-norm regularization is also convex and a global optimal solution can be achieved~\cite{liu2009multi}, thus the following discussions about the sparse learning based feature selection will center around the $\ell_{2,1}$-norm regularization term. The $\ell_{2,1}$-norm regularization also has strong connections with group lasso~\cite{yuan2006model} which will be explained later. By solving the related optimization problem, we can obtain a sparse matrix $\mat{W}$ where many rows are exact zero or of small values, and then the features corresponding to these rows can be eliminated.
\vspace{-0.05in}
\paragraph{Case 4: $p=\infty$, $q=1$} In addition to the $\ell_{2,1}$-norm regularization term, the $\ell_{\infty,1}$-norm regularization is also widely used to achieve joint feature sparsity across multiple targets~\cite{quattoni2009efficient}. In particular, it penalizes the sum of maximum absolute values of each row, such that many rows of the matrix will all be zero.

\vspace{-0.05in}
\subsubsection{Efficient and Robust Feature Selection}
Authors in~\cite{nie2010efficient} propose an efficient and robust feature selection (REFS) method by employing a joint $\ell_{2,1}$-norm minimization on both the loss function and the regularization. Their argument is that the $\ell_{2}$-norm based loss function is sensitive to noisy data while the $\ell_{2,1}$-norm based loss function is more robust to noise. The reason is that $\ell_{2,1}$-norm loss function has a rotational invariant property~\cite{ding2006r}. Consistent with $\ell_{2,1}$-norm regularized feature selection model, a $\ell_{2,1}$-norm regularizer is added to the $\ell_{2,1}$-norm loss function to achieve group feature sparsity. The objective function of REFS is:
\begin{equation}
\small
\min_{\mat{W}}\|\mat{XW}-\mat{Y}\|_{2,1}+\alpha\|\mat{W}\|_{2,1},
\vspace{-0.05in}
\end{equation}
To solve the convex but non-smooth optimization problem, an efficient algorithm is proposed with strict convergence analysis.

It should be noted that the aforementioned REFS is designed for multi-class classification problems where each instance only has one class label. However, data could be associated with multiple labels in many domains such as information retrieval and multimedia annotation. Recently, there is a surge of research work study multi-label feature selection problems by considering label correlations. Most of them, however, are also based on the $\ell_{2,1}$-norm sparse regularization framework~\cite{gu2011correlated,chang2014convex,jian2016multi}.

\vspace{-0.05in}
\subsubsection{Multi-Cluster Feature Selection}
Most of existing sparse learning based approaches build a learning model with the supervision of class labels. The feature selection phase is derived afterwards on the sparse feature coefficients. However, since labeled data is costly and time-consuming to obtain, unsupervised sparse learning based feature selection has received increasing attention in recent years. Multi-Cluster Feature Selection (MCFS)~\cite{cai2010unsupervised} is one of the first attempts. Without class labels to guide the feature selection process, MCFS proposes to select features that can cover multi-cluster structure of the data where spectral analysis is used to measure the correlation between different features.

MCFS consists of three steps. In the first step, it constructs a $p$-nearest neighbor graph to capture the local geometric structure of data and gets the graph affinity matrix $\mat{S}$ and the Laplacian matrix $\mat{L}$. Then a flat embedding that unfolds the data manifold can be obtained by spectral clustering techniques. In the second step, since the embedding of data is known, MCFS takes advantage of them to measure the importance of features by a regression model with a $\ell_{1}$-norm regularization. Specifically, given the $i$-th embedding $\mat{e}_{i}$, MCFS regards it as a regression target to minimize:
\begin{equation}
\small
\min_{w_{i}}\|\mat{X}\mat{w}_{i}-\mat{e}_{i}\|_{2}^{2}+\alpha\|\mat{w}_{i}\|_{1},
\vspace{-0.05in}
\end{equation}
where $\mat{w}_{i}$ denotes the feature coefficient vector for the $i$-th embedding. By solving all $K$ sparse regression problems, MCFS obtains $K$ sparse feature coefficient vectors $\mat{W}=[\mat{w}_{1},...,\mat{w}_{K}]$ and each vector corresponds to one embedding of $\mat{X}$. In the third step, for each feature $f_{j}$, the MCFS score for that feature can be computed as $\text{\emph{MCFS}}(j)=\max_{i}|\mat{W}(j,i)|$. The higher the MCFS score, the more important the feature is.

\vspace{-0.05in}
\subsubsection{$\ell_{2,1}$-norm Regularized Discriminative Feature Selection}
In~\cite{yang2011l2}, the authors propose a new unsupervised feature selection algorithm (UDFS) to select the most discriminative features by exploiting both the discriminative information and feature correlations. First, assume $\tilde{\mat{X}}$ is the centered data matrix such $\tilde{\mat{X}}=\mat{H}_{n}\mat{X}$ and $\mat{G}=[\mat{G}_{1},\mat{G}_{1},...,\mat{G}_{n}]'=\mat{Y}(\mat{Y}'\mat{Y})^{-\frac{1}{2}}$ is the weighted label indicator matrix, where $\mat{H}_{n}=\mat{I}_{n}-\frac{1}{n}\mat{1}_{n}\mat{1}_{n}'$. Instead of using global discriminative information, they propose to utilize the local discriminative information to select discriminative features. The advantage of using local discriminative information are two folds. First, it has been demonstrated to be more important than global discriminative information in many classification and clustering tasks. Second, when it considers the local discriminative information, the data manifold structure is also well preserved. For each data instance $x_{i}$, it constructs a $p$-nearest neighbor set for that instance $\mathcal{N}_{p}(x_{i})=\{x_{i_{1}},x_{i_{2}},...,x_{i_{p}}\}$. Let $\mat{X}_{\mathcal{N}_{p}(i)}=[\mat{x}_{i},\mat{x}_{i_{1}},...,\mat{x}_{i_{p}}]$ denotes the local data matrix around $x_{i}$, then the local total scatter matrix $\mat{S}_{t}^{(i)}$ and local between class scatter matrix $\mat{S}_{b}^{(i)}$ are $\tilde{\mat{X}_{i}}'\tilde{\mat{X}_{i}}$ and $\tilde{\mat{X}_{i}}'\mat{G}_{i}\mat{G}_{i}'\tilde{\mat{X}_{i}}$ respectively, where $\tilde{\mat{X}_{i}}$ is the centered data matrix and $\mat{G}_{(i)}=[\mat{G}_{i},\mat{G}_{i_{1}},...,\mat{G}_{i_{k}}]'$. Note that $\mat{G}_{(i)}$ is a subset from $\mat{G}$ and $\mat{G}_{(i)}$ can be obtained by a selection matrix $\mat{P}_{i}\in\{0,1\}^{n\times (k+1)}$ such that $\mat{G}_{(i)}=\mat{P}_{i}'\mat{G}$. Without label information in unsupervised feature selection, UDFS assumes that there is a linear classifier $\mat{W}\in\mathbb{R}^{d\times s}$ to map each data instance $\mat{x}_{i}\in \mathbb{R}^{d}$ to a low dimensional space $\mat{G}_{i}\in \mathbb{R}^{s}$. Following the definition of global discriminative information~\cite{yang2010image,fukunaga2013introduction}, the local discriminative score for each instance $x_{i}$ is :
\begin{equation}
\small
DS_{i}=tr[(\mat{S}_{t}^{(i)}+\lambda\mat{I}_{d})^{-1}\mat{S}_{b}^{(i)}]=tr[\mat{W}'\mat{X}'\mat{P}_{(i)}\tilde{\mat{X}_{i}}'(\tilde{\mat{X}_{i}}\tilde{\mat{X}_{i}}'+\lambda\mat{I}_{d})^{-1}\tilde{\mat{X}_{i}}\mat{P}_{(i)}'\mat{X}\mat{W}],
\end{equation}
A high local discriminative score indicates that the instance can be well discriminated by $\mat{W}$. Therefore, UDFS tends to train $\mat{W}$ which obtains the highest local discriminative score for all instances in $\mat{X}$; also it incorporates a $\ell_{2,1}$-norm regularizer to achieve feature selection, the objective function is formulated as follows:
\begin{equation}
\small
\min_{\mat{W}'\mat{W}=\mat{I}_{d}}\sum_{i=1}^{n}\{tr[\mat{G}_{(i)}'\mat{H}_{k+1}\mat{G}_{(i)}-DS_{i}]\}+\alpha\|\mat{W}\|_{2,1},
\label{eq:UDFS}
\vspace{-0.05in}
\end{equation}
where $\alpha$ is a regularization parameter to control the sparsity of the learned model.

\vspace{-0.05in}
\subsubsection{Feature Selection Using Nonnegative Spectral Analysis}
Nonnegative Discriminative Feature Selection (NDFS)~\cite{li2012unsupervised} performs spectral clustering and feature selection simultaneously in a joint framework to select a subset of discriminative features. It assumes that pseudo class label indicators can be obtained by spectral clustering techniques. Different from most existing spectral clustering techniques, NDFS imposes nonnegative and orthogonal constraints during the spectral clustering phase. The argument is that with these constraints, the learned pseudo class labels are closer to real cluster results. These nonnegative pseudo class labels then act as regression constraints to guide the feature selection phase. Instead of performing these two tasks separately, NDFS incorporates these two phases into a joint framework.

Similar to the UDFS, we use $\mat{G}=[\mat{G}_{1},\mat{G}_{1},...,\mat{G}_{n}]'=\mat{Y}(\mat{Y}'\mat{Y})^{-\frac{1}{2}}$ to denote the weighted cluster indicator matrix. It is easy to show that we have $\mat{G}\mat{G}'=\mat{I}_{n}$. NDFS adopts a strategy to learn the weight cluster matrix such that the local geometric structure of the data can be well preserved~\cite{shi2000normalized,yu2003multiclass}. The local geometric structure can be preserved by minimizing the normalized graph Laplacian $tr(\mat{G}'\mat{L}\mat{G})$, where $\mat{L}$ is the Laplacian matrix that can be derived from RBF kernel. In addition to that, given the pseudo labels $\mat{G}$, NDFS assumes that there exists a linear transformation matrix $\mat{W}\in\mathbb{R}^{d\times s}$ between the data instances $\mat{X}$ and the pseudo labels $\mat{G}$. These pseudo class labels are utilized as constraints to guide the feature selection process. The combination of these two components results in the following problem:
\begin{equation}
\small
\begin{split}
\min_{\mat{G},\mat{W}}\,tr(\mat{G}'\mat{L}\mat{G})&+\beta(\|\mat{X}\mat{W}-\mat{G}\|_{F}^{2}+\alpha\|\mat{W}\|_{2,1})\\
& \mbox{s.t.} \quad \mat{\mat{G}\mat{G}'=\mat{I}_{n}}, \mat{G}\geq 0,
\end{split}
\label{eq:NDFS}
\end{equation}
where $\alpha$ is a parameter to control the sparsity of the model, and $\beta$ is introduced to balance the contribution of spectral clustering and discriminative feature selection.
\vspace{-0.05in}
\paragraph{\textbf{Discussion}: Sparse learning based feature selection methods have gained increasing popularity in recent years. A merit of such type of methods is that it embeds feature selection into a typical learning algorithm (such as linear regression, SVM, etc.). Thus it can often lead very good performance for the underlying learning algorithm. Also, with sparsity of feature weights, the model poses good interpretability as it enables us to explain why we make such prediction. Nonetheless, there are still some drawbacks of these methods: First, as it directly optimizes a particular learning algorithm by feature selection, the selected features do not necessary achieve good performance in other learning tasks. Second, this kind of methods often involves solving a non-smooth optimization problem, and with complex matrix operations (e.g., multiplication, inverse, etc) in most cases. Hence, the expensive computational cost is another bottleneck}

\vspace{-0.03in}
\subsection{Statistical based Methods}
Another category of feature selection algorithms is based on different statistical measures. As they rely on various statistical measures instead of learning algorithms to assess feature relevance, most of them are filter based methods. In addition, most statistical based algorithms analyze features individually. Hence, feature redundancy is inevitably ignored during the selection phase. We introduce some representative feature selection algorithms in this category.
\vspace{-0.05in}
\subsubsection{Low Variance}
Low Variance eliminates features whose variance are below a predefined threshold. For example, for the features that have the same values for all instances, the variance is 0 and should be removed since it cannot help discriminate instances from different classes. Suppose that the dataset consists of only boolean features, i.e., the feature values are either 0 and 1. As the boolean feature is a Bernoulli random variable, its variance value can be computed as:
\begin{equation}
\small
variance\_score(f_{i})=p(1-p),
\vspace{-0.05in}
\end{equation}
where $p$ denotes the percentage of instances that take the feature value of 1. After the variance of features is obtained, the feature with a variance score below a predefined threshold can be directly pruned.
\vspace{-0.05in}
\subsubsection{T-score}
$T$-score~\cite{davis1986statistics} is used for binary classification problems. For each feature $f_{i}$, suppose that $\mu_{1}$ and $\mu_{2}$ are the mean feature values for the instances from two different classes, $\sigma_{1}$ and $\sigma_{2}$ are the corresponding standard deviations, $n_{1}$ and $n_{2}$ denote the number of instances from these two classes. Then the $t$-score for the feature $f_{i}$ is:
\begin{equation}
\small
t\_score(f_{i})=|\mu_{1}-\mu_{2}|/\sqrt{\frac{\sigma_{1}^{2}}{n_{1}}+\frac{\sigma_{2}^{2}}{n_{2}}}.
\vspace{-0.05in}
\end{equation}
The basic idea of $t$-score is to assess whether the feature makes the means of two classes statistically different, which can be computed as the ratio between the mean difference and the variance of two classes. The higher the $t$-score, the more important the feature is.
\vspace{-0.05in}

\subsubsection{Chi-Square Score}
Chi-square score~\cite{liu1995chi2} utilizes the test of independence to assess whether the feature is independent of the class label. Given a particular feature $f_{i}$ with $r$ different feature values, the Chi-square score of that feature can be computed as:
\begin{equation}
\small
Chi\_square\_score(f_{i})=\sum_{j=1}^{r}\sum_{s=1}^{c}\frac{(n_{js}-\mu_{js})^{2}}{\mu_{js}},
\vspace{-0.05in}
\end{equation}
where $n_{js}$ is the number of instances with the $j$-th feature value given feature $f_{i}$. In addition, $\mu_{js}=\frac{n_{*s}n_{j*}}{n}$, where $n_{j*}$ indicates the number of data instances with the $j$-th feature value given feature $f_{i}$, $n_{*s}$ denotes the number of data instances in class $r$. A higher Chi-square score indicates that the feature is relatively more important.
\vspace{-0.05in}
\subsubsection{Gini Index}
Gini index~\cite{gini1912variability} is also a widely used statistical measure to quantify if the feature is able to separate instances from different classes. Given a feature $f_{i}$ with $r$ different feature values, suppose $\mathcal{W}$ and $\overline{\mathcal{W}}$ denote the set of instances with the feature value smaller or equal to the $j$-th feature value, and larger than the $j$-th feature value, respectively. In other words, the $j$-th feature value can separate the dataset into $\mathcal{W}$ and $\overline{\mathcal{W}}$, then the Gini index score for the feature $f_{i}$ is given as follows:
\begin{equation}
\small
gini\_index\_score(f_{i})=\min_{\mathcal{W}}\left(p(\mathcal{W})(1-\sum_{s=1}^{c}p(C_{s}|\mathcal{W})^{2})+p(\overline{\mathcal{W}})(1-\sum_{s=1}^{c}p(C_{s}|\overline{\mathcal{W}})^{2})\right),
\label{eq:giniIndex}
\vspace{-0.05in}
\end{equation}
where $p(.)$ denotes the probability. For instance, $p(C_{s}|\mathcal{W})$ is the conditional probability of class $s$ given $\mathcal{W}$. For binary classification, Gini Index can take a maximum value of 0.5, it can also be used in multi-class classification problems. Unlike previous statistical measures, the lower the Gini index value, the more relevant the feature is.
\vspace{-0.05in}
\subsubsection{CFS}
The basic idea of CFS~\cite{hall1999feature} is to use a correlation based heuristic to evaluate the worth of a feature subset $\mathcal{S}$:
\begin{equation}
\small
CFS\_score(\mathcal{S})=\frac{k\overline{r_{cf}}}{\sqrt{k+k(k-1)}\overline{r_{ff}}},
\label{eq:cfsScore}
\end{equation}
where the CFS score shows the heuristic ``merit" of the feature subset $\mathcal{S}$ with $k$ features. $\overline{r_{cf}}$ is the mean feature class correlation and $\overline{r_{ff}}$ is the average feature-feature correlation. In Eq.~(\ref{eq:cfsScore}), the numerator indicates the predictive power of the feature set while the denominator shows how much redundancy the feature set has. The basic idea is that a good feature subset should have a strong correlation with class labels and are weakly intercorrelated. To get the feature-class correlation and feature-feature correlation, CFS uses symmetrical uncertainty~\cite{vetterling1992numerical}. As finding the globally optimal subset is computational prohibitive, it adopts a best-search strategy to find a local optimal feature subset. At the very beginning, it computes the utility of each feature by considering both feature-class and feature-feature correlation. It then starts with an empty set and expands the set by the feature with the highest utility until it satisfies some stopping criteria.
\vspace{-0.02in}
\paragraph{\textbf{Discussion}: Most of the statistical based feature selection methods rely on predefined statistical measures to filter out unwanted features, and are simple, straightforward in nature. And the computational costs of these methods are often very low. To this end, they are often used as a preprocessing step before applying other sophisticated feature selection algorithms. Also, as similarity based feature selection methods, these methods often evaluate the importance of features individually and hence cannot handle feature redundancy. Meanwhile, most algorithms in this family can only work on discrete data and conventional data discretization techniques are required to preprocess numerical and continuous variables}
\vspace{-0.05in}
\subsection{Other Methods}
In this subsection, we present other feature selection methods that do not belong to the above four types of feature selection algorithms. In particular, we review hybrid feature selection methods, deep learning based and reconstruction based methods.

Hybrid feature selection methods is a kind of ensemble-based methods that aim to construct a group of feature subsets from different feature selection algorithms, and then produce an aggregated result out of the group. In this way, the instability and perturbation issues of most single feature selection algorithms can be alleviated, and also, the subsequent learning tasks can be enhanced. Similar to conventional ensemble learning methods~\cite{zhou2012ensemble}, hybrid feature selection methods consist of two steps: (1) construct a set of different feature selection results; and (2) aggregate different outputs into a consensus result. Different methods differ in the way how these two steps are performed. For the first step, existing methods either ensemble the selected feature subsets of a single method on different sample subset or ensemble the selected feature subsets from multiple feature selection algorithms. In particular, a sampling method to obtain different sample subsets is necessary for the first case; and typical sampling methods include random sampling and bootstrap sampling. For example,~\cite{saeys2008robust} studied the ensemble feature selection which aggregates a conventional feature selection algorithm such as RELIEF with multiple bootstrapped samples of the training data. In~\cite{abeel2010robust}, the authors improved the stability of SVM-RFE feature selection algorithm by applying multiple random sampling on the original data. The second step involves in aggregating rankings of multiple selected feature subset. Most of the existing methods employ a simple yet effective linear aggregation function~\cite{saeys2008robust,abeel2010robust,yang2011robust}. Nonetheless, other ranking aggregation functions such as Markov chain-based method~\cite{dutkowski2007consensus}, distance synthesis method~\cite{yang2005identifying}, and stacking method~\cite{netzer2009new} are also widely used. In addition to using the aggregation function, another way is to identify the consensus features directly from multiple sample subsets~\cite{loscalzo2009consensus}.

Nowadays, deep learning techniques are popular and successful in various real-world applications, especially in computer vision and natural language processing. Deep learning is distinct from feature selection as deep learning leverages deep neutral networks structures to learn new feature representations while feature selection directly finds relevant features from the original features. From this perspective, the results of feature selection are more human readable and interpretable. Even though deep learning is mainly used for feature learning, there are still some attempts that use deep learning techniques for feature selection. We briefly review these deep learning based feature selection methods. For example, in~\cite{li2015deep}, a deep feature selection model (DFS) is proposed. DFS selects features at the input level of a deep neural network. Typically, it adds a sparse one-to-one linear layer between the input layer and the first hidden layer of a multilayer
perceptrons (MLP). To achieve feature selection, DFS imposes sparse regularization term, then only the features corresponding to nonzero weights are selected. Similarly, in~\cite{roy2015feature}, the authors also propose to select features at the input level of a deep neural network. The difference is that they propose a new concept - net positive contribution, to assess if features are more likely to make the neurons contribute in the classification phase. Since heterogeneous (multi-view) features are prevalent in machine learning and pattern recognition applications,~\cite{zhao2015heterogeneous} proposes to combine deep neural networks with sparse representation for grouped heterogeneous feature selection. It first extracts a new unified representation from each feature group using a multi-modal neural network. Then the importance of features is learned by a kind of sparse group lasso method. In~\cite{wang2014attentional}, the authors propose an attentional neural network, which guides feature selection with cognitive bias. It consists of two modules, a segmentation module, and a classification module. First, given a cognitive bias vector, segmentation module segments out an object belonging to one of classes in the input image. Then, in the classification module, a reconstruction function is applied to the segment to gate the raw image with a threshold for classification. When features are sensitive to a cognitive bias, the cognitive bias will activate the corresponding relevant features.

Recently, data reconstruction error emerged as a new criterion for feature selection, especially for unsupervised feature selection. It defines feature relevance as the capability of features to approximate the original data via a reconstruction function. Among them, Convex Principal Feature Selection (CPFS)~\cite{masaeli2010convex} reformulates the feature selection problem as a convex continuous optimization problem
that minimizes a mean-squared-reconstruction error with linear and sparsity constraint. GreedyFS~\cite{farahat2011efficient} uses a projection matrix to project the original data onto the span of some representative feature vectors and derives an efficient greedy algorithm to obtain these representative features. Zhao et al.~\cite{zhao2016graph} formulates the problem of unsupervised feature selection as the graph regularized data reconstruction. The basic idea is to make the selected features well preserve the data manifold structure of the original data, and reconstruct each data sample via linear reconstruction. A pass-efficient unsupervised feature selection is proposed in~\cite{maung2013pass}. It can be regarded as a modification of the classical pivoted QR algorithm, the basic idea is still to select representative features that can minimize the reconstruction error via linear function. The aforementioned methods mostly use linear reconstruction functions,~\cite{li2017reconstruction} argues that the reconstruction function is not necessarily linear and proposes to learn the reconstruction function automatically function from data. In particular, they define a scheme to embed the reconstruction function learning into feature selection.

\vspace{-0.05in}
\section{Feature Selection with Structured Features}
Existing feature selection methods for conventional data are based on a strong assumption that features are independent of each other (flat) while ignoring the inherent feature structures. However, in many real applications features could exhibit various kinds of structures, e.g., spatial or temporal smoothness, disjoint groups, overlap groups, trees and graphs~\cite{tibshirani2005sparsity,jenatton2011structured,yuan2011efficient,huang2011learning,zhou2012modeling,wang2015multi}. If this is the case, feature selection algorithms incorporating knowledge about the structure information may help find more relevant features and therefore can improve subsequent learning tasks. One motivating example is from bioinformatics, in the study of array CGH, features have some natural spatial order, incorporating such spatial structure can help select more important features and achieve more accurate classification accuracy. Therefore, in this section, we discuss some representative feature selection algorithms which explicitly consider feature structures. Specifically, we will focus on group structure, tree structure and graph structure.

A popular and successful approach to achieve feature selection with structured features is to minimize an empirical error penalized by a structural regularization term:
\begin{equation}
\small
\mat{w}=\argmin_{\mat{w}}\, loss(\mat{w};\mat{X},\mat{y})+\alpha \, penalty(\mat{w},\mathcal{G}),
\label{eq:fs_structural}
\vspace{-0.05in}
\end{equation}
where $\mathcal{G}$ denotes the structures among features and $\alpha$ is a trade-off parameter between the loss function and the structural regularization term.
To achieve feature selection, $penalty(\mat{w},\mathcal{G})$ is usually set to be a sparse regularization term. Note that the above formulation is similar to that in Eq.~(\ref{eq:fs_regularization}), the only difference is that for feature selection with structured features, we explicitly consider the structural information $\mathcal{G}$ among features in the sparse regularization term.

\vspace{-0.05in}
\subsection{Feature Selection with Group Feature Structures}
\begin{figure}[!htbp]
  \centering
    \includegraphics[width=0.6\textwidth]{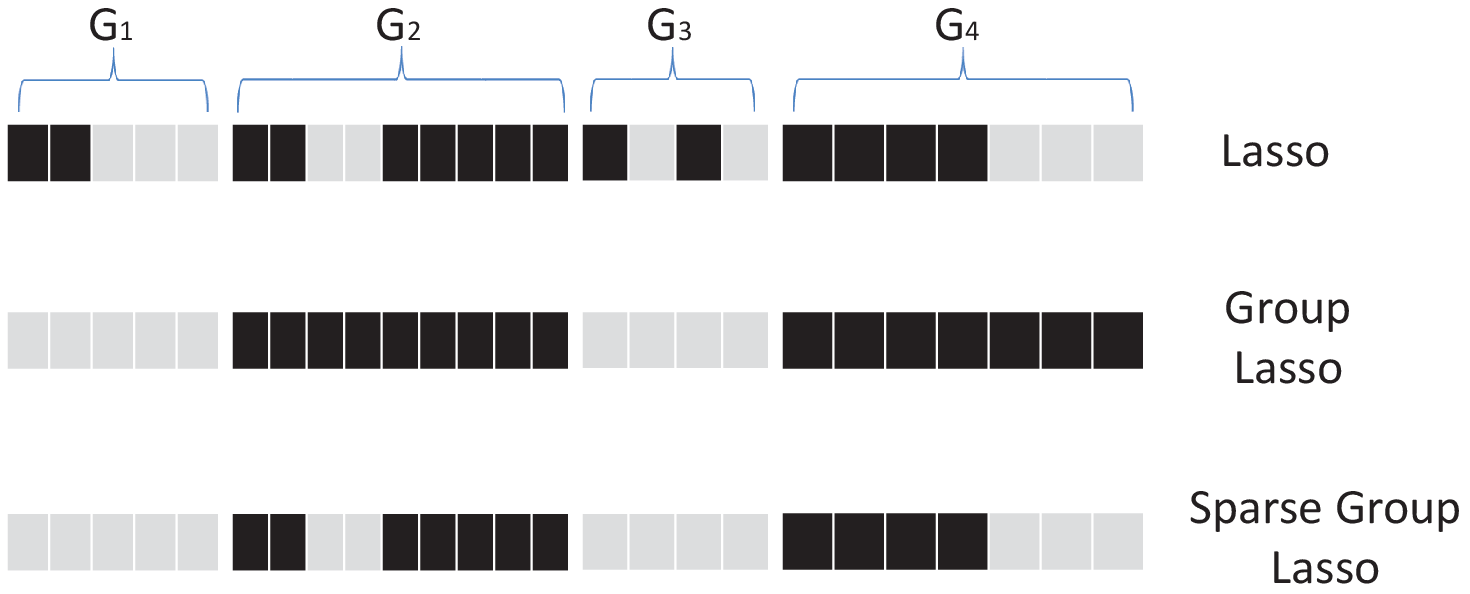}
      \caption{Illustration of Lasso, Group Lasso, and Sparse Group Lasso. The feature set can be divided into four groups $G_{1}$, $G_{2}$, $G_{3}$ and $G_{4}$. The column with dark color denotes selected features while the column with light color denotes unselected features.}
\label{fig:GroupLasso}
\vspace{-0.18in}
\end{figure}

First, features could exhibit group structures. One of the most common examples is that in multifactor analysis-of-variance (ANOVA), each factor is associated with several groups and can be expressed by a set of dummy features~\cite{yuan2006model}. Some other examples include different frequency bands represented as groups in signal processing~\cite{mcauley2005subband} and genes with similar functionalities acting as groups in bioinformatics~\cite{ma2007supervised}. Therefore, when performing feature selection, it is more appealing to model the group structure explicitly.

\vspace{-0.05in}
\subsubsection{Group Lasso}
Group Lasso~\cite{yuan2006model,bach2008consistency,jacob2009group,meier2008group}, which derives feature coefficients from certain groups to be small or exact zero, is a solution to this problem. In other words, it selects or ignores a group of features as a whole. The difference between Lasso and Group Lasso is shown by the illustrative example in Fig.~\ref{fig:GroupLasso}. Suppose that these features come from 4 different groups and there is no overlap between these groups. Lasso completely ignores the group structures among features, and the selected features are from four different groups. On the contrary, Group Lasso tends to select or not select features from different groups as a whole. As shown in the figure, Group Lasso only selects the second and the fourth group $G_{2}$ and $G_{4}$, features in the other two groups $G_{1}$ and $G_{3}$ are not selected. Mathematically, Group Lasso first uses a $\ell_{2}$-norm regularization term for feature coefficients $\mat{w}_{i}$ in each group $G_{i}$, then it performs a $\ell_{1}$-norm regularization for all previous $\ell_{2}$-norm terms. The objective function of Group Lasso is formulated as follows:
\begin{equation}
\small
\min_{\mat{w}}\, loss(\mat{w};\mat{X},\mat{y})+\alpha \, \sum_{i=1}^{g}h_{i}\|\mat{w}_{G_{i}}\|_{2},
\label{eq:fs_grouplasso}
\vspace{-0.05in}
\end{equation}
where $h_{i}$ is a weight for the $i$-th group $\mat{w}_{G_{i}}$ which can be considered as a prior to measuring the contribution of the $i$-th group in the feature selection process.

\vspace{-0.05in}
\subsubsection{Sparse Group Lasso}
Once Group Lasso selects a group, all the features in the selected group will be kept. However, in many cases, not all features in the selected group could be useful, and it is desirable to consider the intrinsic feature structures and select features from different selected groups simultaneously (as illustrated in Fig.~\ref{fig:GroupLasso}). Sparse Group Lasso~\cite{friedman2010note,peng2010regularized} takes advantage of both Lasso and Group Lasso, and it produces a solution with simultaneous intra-group and inter-group sparsity. The sparse regularization term of Sparse Group Lasso is a combination of the penalty term of Lasso and Group Lasso:
\begin{equation}
\small
\min_{\mat{w}}\, loss(\mat{w};\mat{X},\mat{y})+\alpha\|\mat{w}\|_{1}+(1-\alpha) \, \sum_{i=1}^{g}h_{i}\|\mat{w}_{G_{i}}\|_{2},
\label{eq:fs_grouplasso}
\vspace{-0.05in}
\end{equation}
where $\alpha$ is parameter between 0 and 1 to balance the contribution of inter-group sparsity and intra-group sparsity for feature selection. The difference between Lasso, Group Lasso and Sparse Group Lasso is shown in Fig.~\ref{fig:GroupLasso}.

\subsubsection{Overlapping Sparse Group Lasso}
Above methods consider the disjoint group structures among features. However, groups may also overlap with each other~\cite{jacob2009group,jenatton2011structured,zhao2009composite}. One motivating example is the usage of biologically meaningful gene/protein groups mentioned in~\cite{ye2012sparse}. Different groups of genes may overlap, i.e., one protein/gene may belong to multiple groups. A general Overlapping Sparse Group Lasso regularization is similar to the regularization term of Sparse Group Lasso. The difference is that different feature groups $G_{i}$ can have an overlap, i.e., there exist at least two groups $G_{i}$ and $G_{j}$ such that $G_{i}\bigcap G_{j}\neq \emptyset$.

\subsection{Feature Selection with Tree Feature Structures}
In addition to the group structures, features can also exhibit tree structures. For example, in face recognition, different pixels can be represented as a tree, where the root node indicates the whole face, its child nodes can be different organs, and each specific pixel is considered as a leaf node. Another motivating example is that genes/proteins may form certain hierarchical tree structures~\cite{liu2010moreau}. Recently, Tree-guided Group Lasso is proposed to handle the feature selection for features that can be represented in an index tree~\cite{kim2010tree,liu2010moreau,jenatton2010proximal}.
\begin{figure}[!htbp]
  \centering
    \includegraphics[width=0.35\textwidth]{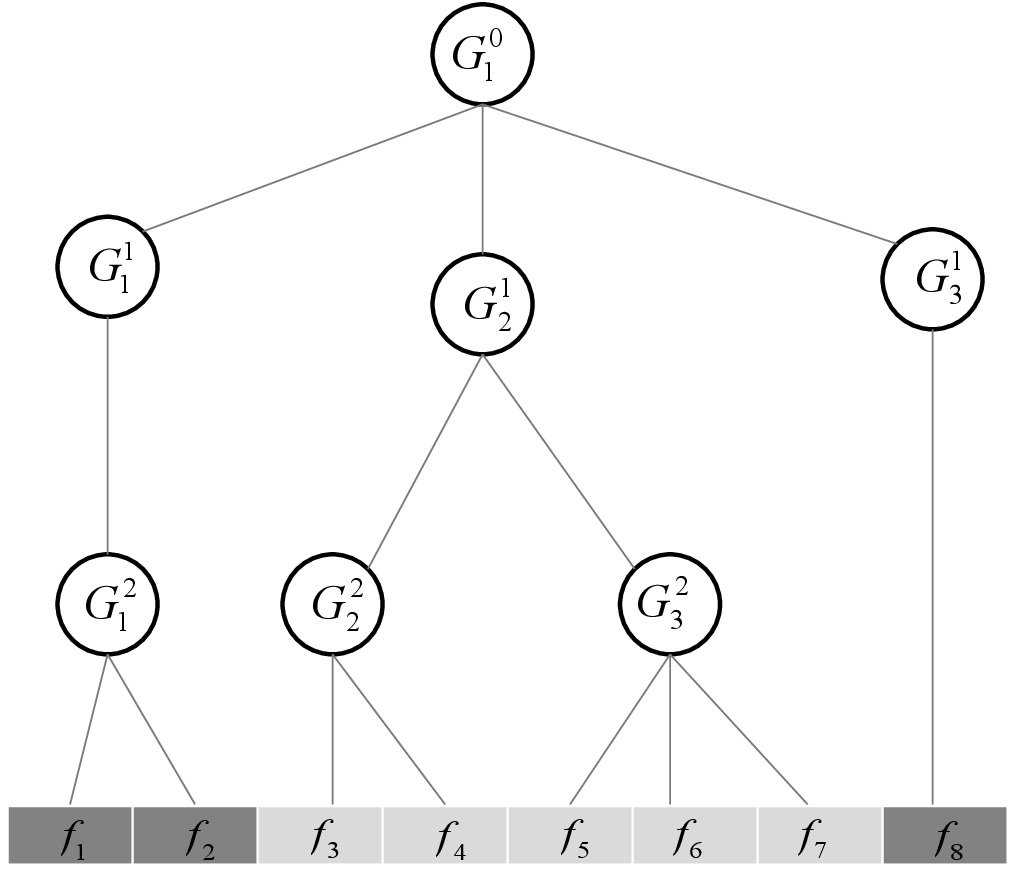}
      \caption{Illustration of the tree structure among features. These eight features form a simple index tree with a depth of 3.}
\vspace{-0.2in}
\label{fig:TreeLasso}
\end{figure}

\vspace{-0.05in}
\subsubsection{Tree-guided Group Lasso}
In Tree-guided Group Lasso~\cite{liu2010moreau}, the structure over the features can be represented as a tree with leaf nodes as features. Each internal node denotes a group of features such that the internal node is considered as a root of a subtree and the group of features is considered as leaf nodes. Each internal node in the tree is associated with a weight that represents the height of its subtree, or how tightly the features in this subtree are correlated.

In Tree-guided Group Lasso, for an index tree $\mathcal{G}$ with a depth of $d$, $\mathcal{G}_{i}=\{G_{1}^{i},G_{2}^{i},...,G_{n_{i}}^{i}\}$ denotes the whole set of nodes (features) in the $i$-th level (the root node is in level 0), and $n_{i}$ denotes the number of nodes in the level $i$. Nodes in Tree-guided Group Lasso have to satisfy the following two conditions: (1) internal nodes from the same depth level have non-overlapping indices, i.e., $G_{j}^{i}\bigcap G_{k}^{i}=\emptyset$, $\forall i=1,2,...,d$, $j\neq k$, $i\leq j, k\leq n_{i}$; and (2) if $G_{m}^{i-1}$ is the parent node of $G_{j}^{i}$, then $G_{j}^{i}\subseteq G_{m}^{i-1}$.

We explain these conditions via an illustrative example in Fig.~\ref{fig:TreeLasso}. In the figure, we can observe that 8 features are organized in an indexed tree of depth 3. For the internal nodes in each level, we have $G_{1}^{0}=\{f_{1},f_{2},f_{3},f_{4},f_{5},f_{6},f_{7},f_{8}\}$, $G_{1}^{1}=\{f_{1},f_{2}\},G_{2}^{1}=\{f_{3},f_{4},f_{5},f_{6},f_{7}\},G_{3}^{1}=\{f_{8}\}$, $G_{1}^{2}=\{f_{1},f_{2}\},G_{2}^{2}=\{f_{3},f_{4}\},G_{3}^{2}=\{f_{5},f_{6},f_{7}\}$. $G_{1}^{0}$ is the root node of the index tree. In addition, internal nodes from the same level do not overlap while the parent node and the child node have some overlap such that the features of the child node is a subset of those of the parent node. In this way, the objective function of Tree-guided Group Lasso is:
\begin{equation}
\small
\min_{\mat{w}}\, loss(\mat{w};\mat{X},\mat{y})+\alpha\, \sum_{i=0}^{d}\sum_{j=1}^{n_{i}}h_{j}^{i}\|\mat{w}_{G_{j}^{i}}\|_{2},
\label{eq:fs_treelasso}
\vspace{-0.05in}
\end{equation}
where $\alpha\geq 0$ is a regularization parameter and $h_{j}^{i}\geq 0$ is a predefined parameter to measure the contribution of the internal node $G_{j}^{i}$. Since parent node is a superset of its child nodes, thus, if a parent node is not selected, all of its child nodes will not be selected. For example, as illustrated in Fig.~\ref{fig:TreeLasso}, if the internal node $G_{2}^{1}$ is not selected, both of its child nodes $G_{2}^{2}$ and $G_{3}^{2}$ will not be selected.

\vspace{-0.05in}
\subsection{Feature Selection with Graph Feature Structures}
In many cases, features may have strong pairwise interactions. For example, in natural language processing, if we take each word as a feature, we have synonyms and antonyms relationships between different words~\cite{fellbaum1998wordnet}. Moreover, many biological studies show that there exist strong pairwise dependencies between genes. Since features show certain kinds of dependencies in these cases, we can model them by an undirected graph, where nodes represent features and edges among nodes show the pairwise dependencies between features~\cite{sandler2009regularized,kim2009statistical,yang2012feature}. We can use an undirected graph $\mathcal{G}(N,E)$ to encode these dependencies. Assume that there are $n$ nodes $N=\{N_{1},N_{2},...,N_{n}\}$ and a set of $e$ edges $\{E_{1},E_{2},...,E_{e}\}$ in $\mathcal{G}(N,E)$. Then node $N_{i}$ corresponds to the $i$-th feature and the pairwise feature dependencies can be represented by an adjacency matrix $\mat{A}\in\mathbb{R}^{N_{n}\times N_{n}}$.

\vspace{-0.05in}
\subsubsection{Graph Lasso}
Since features exhibit graph structures, when two nodes (features) $N_{i}$ and $N_{j}$ are connected by an edge in $\mathcal{G}(N,E)$, the features $f_{i}$ and $f_{j}$ are more likely to be selected together, and they should have similar feature coefficients. One way to achieve this target is via Graph Lasso -- adding a graph regularizer for the feature graph on the basis of Lasso~\cite{ye2012sparse}. The formulation is:
\begin{equation}
\small
\min_{\mat{w}}\, loss(\mat{w};\mat{X},\mat{y})+\alpha\|\mat{w}\|_{1}+(1-\alpha)\, \sum_{i,j}\mat{A}(i,j)(\mat{w}_{i}-\mat{w}_{j})^{2},
\label{eq:fs_graphlasso}
\vspace{-0.05in}
\end{equation}
where the first regularization term $\alpha\|\mat{w}\|_{1}$ is from Lasso while the second term ensures that if a pair of features show strong dependency, i.e., large $\mat{A}(i,j)$, their feature coefficients should also be similar to each other.

\vspace{-0.05in}
\subsubsection{GFLasso}
In Eq.~(\ref{eq:fs_graphlasso}), Graph Lasso encourages features connected together have similar feature coefficients. However, features can also be negatively correlated. In this case, the feature graph $\mathcal{G}(N,E)$ is represented by a signed graph, with both positive and negative edges. GFLasso~\cite{kim2009statistical} is proposed to model both positive and negative feature correlations, the objective function is:
\begin{equation}
\small
\min_{\mat{w}}\, loss(\mat{w};\mat{X},\mat{y})+\alpha\|\mat{w}\|_{1}+(1-\alpha)\, \sum_{i,j}\mat{A}(i,j)|\mat{w}_{i}-\mbox{sign}(r_{i,j})\mat{w}_{j}|,
\label{eq:fs_gflasso}
\vspace{-0.05in}
\end{equation}
where $r_{i,j}$ indicates the correlation between two features $f_{i}$ and $f_{j}$. When two features are positively correlated, we have $\mat{A}(i,j)=1$ and $r_{i,j}>0$, and the penalty term forces the feature coefficients $\mat{w}_{i}$ and $\mat{w}_{j}$ to be similar; on the other hand, if two features are negatively correlated, we have $\mat{A}(i,j)=1$ and $r_{i,j}<0$, and the penalty term makes the feature coefficients $\mat{w}_{i}$ and $\mat{w}_{j}$ to be dissimilar. A major limitation of GFLasso is that it uses pairwise sample correlations to measure feature dependencies, which may lead to additional estimation bias. The feature dependencies cannot be correctly estimated when the sample size is small.

\vspace{-0.05in}
\subsubsection{GOSCAR}
To address the limitations of GFLasso,~\cite{yang2012feature} propose GOSCAR by putting a $\ell_{\infty}$-norm regularization to enforce pairwise feature coefficients to be equivalent if two features are connected in the feature graph. The formulation is:
\begin{equation}
\small
\min_{\mat{w}}\, loss(\mat{w};\mat{X},\mat{y})+\alpha\|\mat{w}\|_{1}+(1-\alpha)\, \sum_{i,j}\mat{A}(i,j)\mbox{max}(|\mat{w}_{i}|,|\mat{w}_{j}|).
\label{eq:fs_goscar}
\end{equation}
In the above formulation, the $\ell_{1}$-norm regularization is used for feature selection while the pairwise $\ell_{\infty}$-norm term penalizes large coefficients. The pairwise $\ell_{\infty}$-norm term can be decomposed as $\mbox{max}(|\mat{w}_{i}|,|\mat{w}_{j}|)=\frac{1}{2}(|\mat{w}_{i}+\mat{w}_{j}|+|\mat{w}_{i}-\mat{w}_{j}|)=|\mat{u}'\mat{w}|+|\mat{v}'\mat{w}|$, where $\mat{u}$ and $\mat{v}$ are sparse vectors with only two nonzero entries such that $\mat{u}_{i}=\mat{u}_{j}=\frac{1}{2}$, $\mat{v}_{i}=-\mat{v}_{j}=\frac{1}{2}$.
\vspace{-0.02in}
\paragraph{\textbf{Discussion}: This family of algorithms explicitly take the structures among features as prior knowledge and feed into feature selection. Therefore, the selected features could enhance subsequent learning tasks. However, most of these methods are based on the sparse learning framework, and often involves in solving complex optimization algorithms. Thus, computational costs could be relatively high. Moreover, the feature structure are often given a priori, it is still a challenging problem to automatically infer the structures from data for feature selection}

\vspace{-0.05in}
\section{Feature Selection with Heterogeneous Data}
Traditional feature selection algorithms are heavily based on the data i.i.d. assumption. However, heterogeneous data from different sources is becoming more and more prevalent in the era of big data. For example, in the medical domain, genes are often associated with different types of clinical features. Since data of each source can be noisy, partial, or redundant, how to find relevant sources and how to fuse them together for effective feature selection is a challenging problem. Another example is in social media platforms, instances of high dimensionality are often linked together, how to integrate link information to guide feature selection is another difficult problem. In this section, we review current feature selection algorithms for heterogeneous data from three aspects: (1) feature selection for linked data; (2) multi-source feature selection; and (3) multi-view feature selection. Note that multi-source and multi-view feature selection are different in two ways: First, multi-source feature selection aims to select features from the original feature space by integrating multiple sources while multi-view feature selection selects features from different feature spaces for all views simultaneously. Second, multi-source feature selection normally ignores the correlations among sources while multi-view feature selection exploits relations among features from different sources.
\vspace{-0.05in}
\subsection{Feature Selection Algorithms with Linked Data}
Linked data is ubiquitous in real-world applications such as Twitter (tweets linked by hyperlinks), Facebook (users connected by friendships) and biological systems (protein interactions). Due to different types of links, they are distinct from traditional attribute-value data (or so-called ``flat" data).

\begin{figure}[!t]
\centering
\begin{minipage}{0.4\textwidth}
\centering
\subfigure[Linked Data\label{fig:linkedfeature-a}]
{\includegraphics[width=\textwidth]{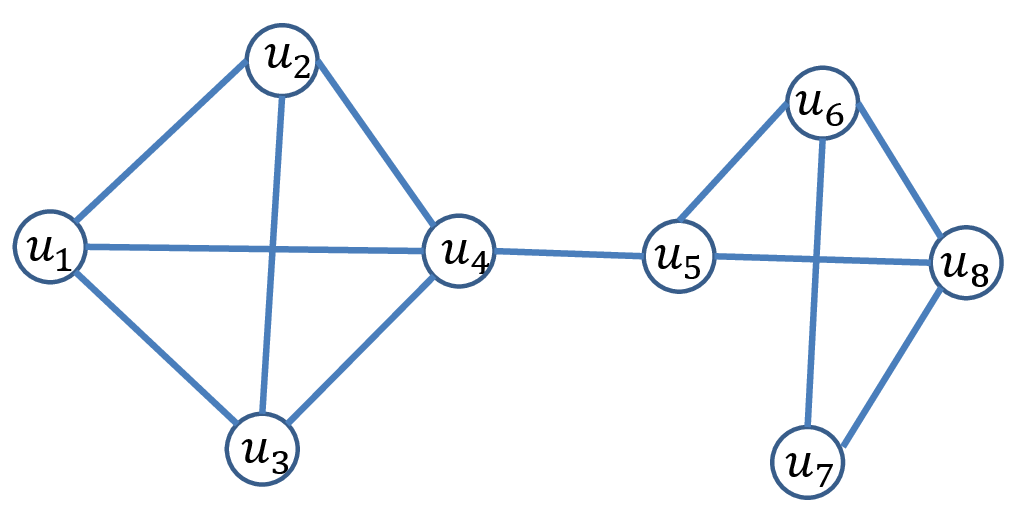}}
\end{minipage}
\begin{minipage}{0.45\textwidth}
\centering
\subfigure[Linked Data Representation\label{fig:linkedfeature-c}]
{\includegraphics[width=\textwidth]{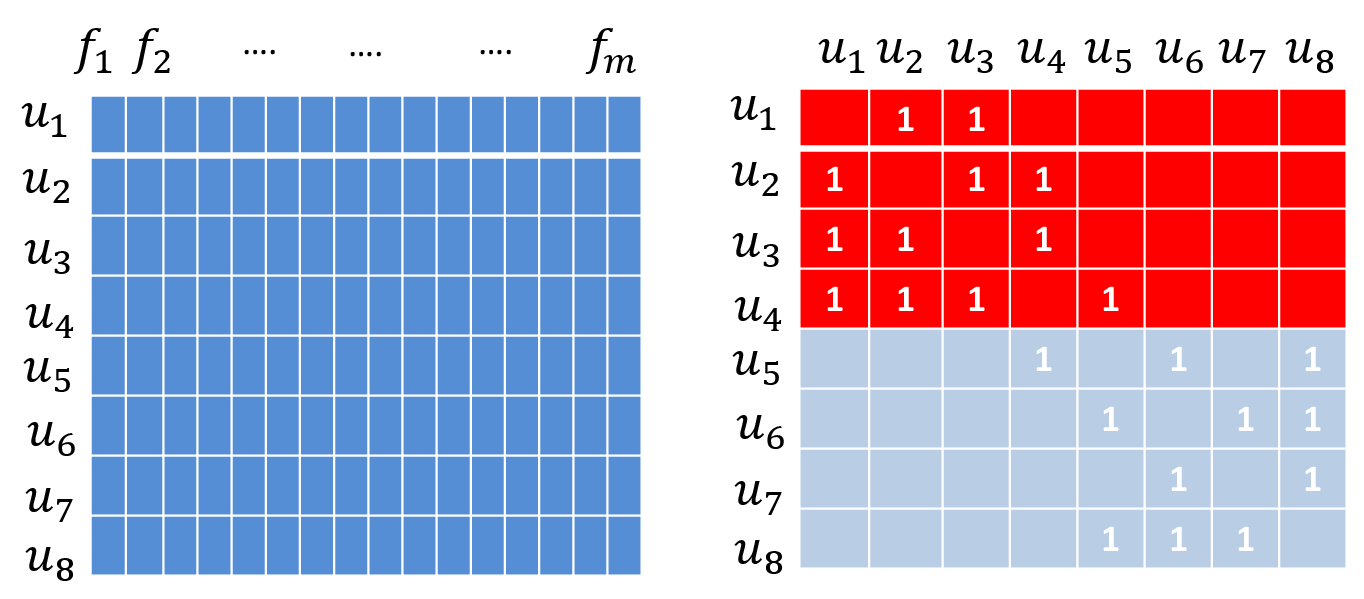}}
\end{minipage}
\centering
\caption{An illustrative example of linked data.}
\vspace{-0.1in}
\label{fig:linkedfeature}
\end{figure}

Fig.~\ref{fig:linkedfeature} illustrates an example of linked data and its representation. Fig.~\ref{fig:linkedfeature-a} shows 8 linked instances, the feature information is illustrated in the left part of Fig.~\ref{fig:linkedfeature-c}. Linked data provides an extra source of information, which can be represented by an adjacency matrix, illustrated in the right part of Fig.~\ref{fig:linkedfeature-c}. Many linked data related learning tasks are proposed such as collective classification~\cite{macskassy2007classification,sen2008collective}, relational learning~\cite{long2006spectral,long2007probabilistic,li2017toward}, link prediction~\cite{liben2007link,backstrom2011supervised,chen2016fascinate}, and active learning~\cite{bilgic2010active,hu2013actnet}, but the task of feature selection is not well studied due to some of its unique challenges: (1) how to exploit relations among data instances; (2) how to take advantage of these relations for feature selection; and (3) linked data is often unlabeled, how to evaluate the relevance of features without labels. Recent years have witnessed a surge of research interests in performing feature selection on linked data~\cite{gu2011towards,tang2012feature,tang2012unsupervised,tang2013coselect,wei2015efficient,wei2016unsupervisedaaai,li2015unsupervised,li2016robust,li2016toward,cheng2017unsupervised}. Next, we introduce some representative algorithms in this family.

\vspace{-0.05in}
\subsubsection{Feature Selection on Networks}
In~\cite{gu2011towards}, the authors propose a supervised feature selection algorithm (FSNet) based on Laplacian Regularized Least Squares (LapRLS). In detail, they propose to use a linear classifier to capture the relationship between content information and class labels, and incorporate link information by graph regularization. Suppose that $\mat{X}\in\mathbb{R}^{n\times d}$ denotes the content matrix and $\mat{Y}\in\mathbb{R}^{n\times c}$ denotes the one-hot label matrix, $\mat{A}$ denotes the adjacency matrix for all $n$ linked instances. FSNet first attempts to learn a linear classifier $\mat{W}\in\mathbb{R}^{d\times c}$ to map $\mat{X}$ to $\mat{Y}$:
\begin{equation}
\small
\min_{\mat{W}}\|\mat{XW}-\mat{Y}\|_{F}^{2}+\alpha\|\mat{W}\|_{2,1}+\beta\|\mat{W}\|_{F}^{2}.
\vspace{-0.05in}
\end{equation}
The term $\|\mat{W}\|_{2,1}$ is included to achieve joint feature sparsity across different classes. $\|\mat{W}\|_{F}^{2}$ prevents the overfitting of the model. To capture the correlation between link information and content information to select more relevant features, FSNet uses the graph regularization and the basic assumption is that if two instances are linked, their class labels are likely to be similar, which results in the following objective function:
\begin{equation}
\small
\min_{\mat{W}}\|\mat{XW}-\mat{Y}\|_{F}^{2}+\alpha\|\mat{W}\|_{2,1}+\beta\|\mat{W}\|_{F}^{2}+\gamma tr(\mat{W}'\mat{X}'\mat{L}\mat{XW}),
\label{eq:FSNet}
\vspace{-0.05in}
\end{equation}
where $tr(\mat{W}'\mat{X}'\mat{L}\mat{XW})$ is the graph regularization, and $\gamma$ balances the contribution of content information and link information for feature selection.

\vspace{-0.05in}
\subsubsection{Feature Selection for Social Media Data (LinkedFS)}
~\cite{tang2012feature} investigate the feature selection problem on social media data by evaluating various social relations such as CoPost, CoFollowing, CoFollowed, and Following. These four types of relations are supported by social correlation theories such as homophily~\cite{mcpherson2001birds} and social influence~\cite{marsden1993network}. We use the CoPost relation as an example to illustrate how these relations can be integrated into feature selection. Let $\mat{p}=\{p_{1},p_{2},...,p_{N}\}$ be the post set and $\mat{X}\in \mathbb{R}^{N\times d}$ be the matrix representation of these posts; $\mat{Y}\in\mathbb{R}^{n\times c}$ denotes the label matrix; $\mat{u}=\{u_{1},u_{2},...,u_{n}\}$ denotes the set of $n$ users and their link information is encoded in an adjacency matrix $\mat{A}$; $\mat{P}\in\mathbb{R}^{n\times N}$ denotes the user-post relationships such that $\mat{P}(i,j)=1$ if $u_{i}$ posts $p_{j}$, otherwise 0. To integrate the CoPost relations among users into the feature selection framework, the authors propose to add a regularization term to enforce the hypothesis that the class labels (i.e., topics) of posts by the same user are similar, resulting in the following objective function:
\begin{equation}
\small
\min_{\mat{W}}\|\mat{XW}-\mat{Y}\|_{F}^{2}+\alpha\|\mat{W}\|_{2,1}+\beta\sum_{u\in\mat{u}}\sum_{\{p_{i},p_{j}\}\in\mat{p}_{u}}\|\mat{X}(i,:)\mat{W}-\mat{X}(j,:)\mat{W}\|_{2}^{2},
\label{eq:linkedfscopost}
\vspace{-0.05in}
\end{equation}
where $\mat{p}_{u}$ denotes the set of posts by user $u$. The parameter $\alpha$ controls the sparsity of $\mat{W}$ in rows across all class labels and $\beta$ controls the contribution of the CoPost relations.
\vspace{-0.05in}
\subsubsection{Unsupervised Feature Selection for Linked Data}
Linked Unsupervised Feature Selection (LUFS)~\cite{tang2012unsupervised} is an unsupervised feature selection framework for linked data. Without label information to assess feature relevance, LUFS assumes the existence of pseudo labels, and uses $\mat{Y}\in\mathbb{R}^{n\times c}$ to denote the pseudo label matrix such that each row of $\mat{Y}$ has only one nonzero entry. Also, LUFS assumes a linear mapping matrix $\mat{W}\in\mathbb{R}^{d\times c}$ between feature $\mat{X}$ and $\mat{Y}$. First, to consider the constraints from link information, LUFS employs social dimension approach~\cite{tang2009relational} to obtain the hidden factors $\mat{H}$ that incur the interdependency among instances. Then, according to the Linear Discriminative Analysis, within, between and total hidden factor scatter matrix $\mat{S}_{w}$, $\mat{S}_{b}$ and $\mat{S}_{t}$ are defined as $\mat{S}_{w}=\mat{Y}'\mat{Y}-\mat{Y}'\mat{F}\mat{F}'\mat{Y}$, $\mat{S}_{b}=\mat{Y}'\mat{F}\mat{F}'\mat{Y}$, $\mat{S}_{t}=\mat{Y}'\mat{Y}$ respectively, where $\mat{F}=\mat{H}(\mat{H}'\mat{H})^{-\frac{1}{2}}$ is the weighted hidden factor matrix. Considering the fact that instances with similar hidden factors are similar and instances with different hidden factors are dissimilar, the constraint from link information can be incorporated by maximizing $tr((\mat{S}_{t})^{-1}\mat{S}_{b})$. Second, to take advantage of feature information, LUFS obtains the constraints by spectral analysis to minimize $tr(\mat{Y}'\mat{L}\mat{Y})$,
where $\mat{L}$ is the Laplacian matrix derived from feature affinity matrix $\mat{S}$. With these, the objective function of LUFS is formulated as follows:
\begin{equation}
\small
\min_{W}tr(\mat{Y}'\mat{L}\mat{Y})-\alpha tr((\mat{S}_{t})^{-1}\mat{S}_{b}),
\label{eq:LUFS-obj1}
\vspace{-0.05in}
\end{equation}
where $\alpha$ is a regularization parameter to balance the contribution from these two constraints. To achieve feature selection, LUFS further adds a $\ell_{2,1}$-norm regularization term on $\mat{W}$, and with spectral relaxation of the pseudo-class label matrix, the objective function in Eq.~(\ref{eq:LUFS-obj1}) can be eventually represented as:
\begin{equation}
\small
\begin{split}
\min_{\mat{W}}tr(\mat{W}'(\mat{X}'\mat{LX}&+\alpha\mat{X}'(\mat{I}_{n}-\mat{F}\mat{F}'))\mat{W})+\beta\|\mat{W}\|_{2,1}\\
\mbox{s.t.}& \quad \mat{W}'(\mat{X}'\mat{X}+\lambda\mat{I}_{d})\mat{W}=\mat{I}_{c},
\end{split}
\vspace{-0.05in}
\end{equation}
where $\beta$ controls the sparsity of $\mat{W}$ in rows and $\lambda\mat{I}_{d}$ makes $\mat{X}'\mat{X}+\lambda\mat{I}_{d}$ invertible.

\subsubsection{Robust Unsupervised Feature Selection for Networked Data}
LUFS performs network structure modeling and feature selection separately, and the feature selection heavily depends on the quality of extracted latent representations. In other words, the performance of LUFS will be jeopardized when there are a lot of noisy links in the network. ~\cite{li2016robust} propose a robust unsupervised feature selection framework (NetFS) to embed latent representation learning into feature selection. Specifically, let $\mat{X}\in\mathbb{R}^{n\times d}$ and $\mat{A}\in\mathbb{R}^{n\times n}$ denote the feature matrix and adjacency matrix respectively. NetFS first uncovers a low-rank latent representation $\mat{U}$ by a symmetric NMF model. The latent representation describes a set of diverse affiliation factors hidden in a network, and instances with similar latent representations are more likely to be connected to each other than the instances with dissimilar latent representations. As latent factors encode some hidden attributes of instances, they should be related to some features. Thus, NetFS takes $\mat{U}$ as a constraint to perform feature selection via:
\begin{equation}
\small
\min_{\mat{U}\geq 0, \mat{W}}\|\mat{XW}-\mat{U}\|_{F}^{2}+\alpha\|\mat{W}\|_{2,1}+\frac{\beta}{2}\|\mat{A}-\mat{U}\mat{U}'\|_{F}^{2},
\end{equation}
where $\alpha$ and $\beta$ are two balance parameters. By embedding latent representation learning into feature selection, these two phases could help and boost each other. Feature information can help learn better latent representations which are robust to noisy links, and better latent representations can fill the gap of limited label information and rich link information to guide feature selection. The authors further extended the NetFS model to the dynamic case to obtain a subset of relevant features continuously when both the feature information and network structure evolve over time~\cite{li2016toward}. In addition to positive links, many real-world networks also contain negative links, such as distrust relations in Epinions and foes in Slashdot. Based on NetFS, the authors in~\cite{cheng2017unsupervised} further study if negative links have added value over positive links in finding more relevant features.

\subsection{Multi-Source Feature Selection}
For many learning tasks, we often have multiple data sources for the same set of data instances. For example, recent advancements in bioinformatics reveal that non-coding RNA species function across a variety of biological process. The task of multi-source feature selection in this case is formulated as follows: given $m$ sources of data depicting the same set of $n$ instances, and their matrix representations $\mat{X}_{1}\in\mathbb{R}^{n\times d_{1}},\mat{X}_{2}\in\mathbb{R}^{n\times d_{2}},...,\mat{X}_{m}\in\mathbb{R}^{n\times d_{m}}$ (where $d_{1},...,d_{m}$ denote the feature dimensions), select a subset of relevant features from a target source (e.g., $\mat{X}_{i}$) by taking advantage of all information from $m$ sources.

\vspace{-0.05in}
\subsubsection{Multi-Source Feature Selection via Geometry-Dependent Covariance Analysis (GDCOV)}
To integrate information from multiple sources, ~\cite{zhao2008multi} propose an intuitive way to learn a global geometric pattern from all sources that reflects the intrinsic relationships among instances~\cite{lanckriet2004learning}. They introduce a concept of geometry-dependent covariance that enables the usage of the global geometric pattern in covariance analysis for feature selection. Given multiple
local geometric patterns in multiple affinity matrices $\mat{S}_{i}$, where $i$ denotes the $i$-th data source, a global pattern can be obtained by linearly combining all affinity matrices as $\mat{S}=\sum_{i=1}^{m}\alpha_{i}\mat{S}_{i}$, where $\alpha_{i}$ controls the contribution of the $i$-th source. With the global geometric pattern obtained from multiple data sources, one can build a geometry-dependent sample covariance matrix for the target source $\mat{X}_{i}$ as $\mat{C}=\frac{1}{n-1}\mat{\Pi}\mat{X}_{i}'(\mat{S}-\frac{\mat{S}\mat{1}\mat{1}'\mat{S}}{\mat{1}'\mat{S}\mat{1}})\mat{X}_{i}\mat{\Pi}$, where $\mat{\Pi}$ is a diagonal matrix with $\mat{\Pi}(j,j)=\|\mat{D}^{\frac{1}{2}}\mat{X}_{i}(:,j)\|^{-1}$, and $\mat{D}$ is also a diagonal matrix from $\mat{S}$ with $\mat{D}(k,k)=\sum_{j=1}^{n}\mat{S}(k,j)$.

After getting a geometry-dependent sample covariance matrix, a subsequent question is how to use it effectively for feature selection. Basically, two methods are proposed. The first method, GPCOVvar sorts the diagonal of the covariance matrix and selects the features that have the highest variances. Selecting features based on this approach is equivalent to choosing features that are consistent with the global geometry pattern. The second method, GPCOVspca, applies Sparse Principal Component Analysis (SPCA)~\cite{d2007direct} to select features that can retain the total variance maximally. Hence, it considers interactions among features and can select features with less redundancy.

\vspace{-0.05in}
\subsection{Feature Selection Algorithms with Multi-View Data}
\begin{figure}[!t]
\centering
\begin{minipage}{0.35\textwidth}
\centering
\subfigure[Multi-Source Feature Selection\label{fig:Multi-source}]
{\includegraphics[width=\textwidth]{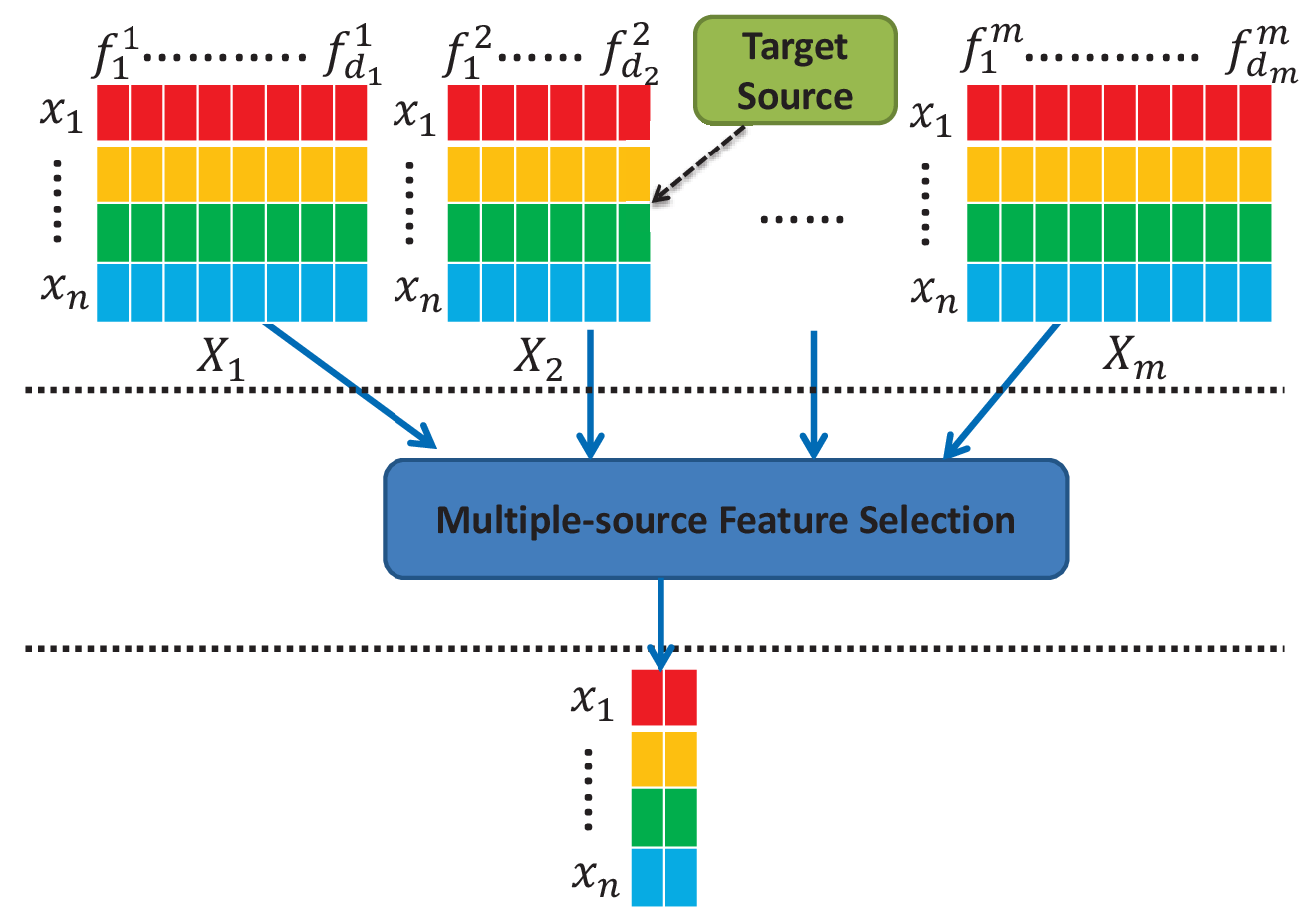}}
\end{minipage}
\begin{minipage}{0.35\textwidth}
\centering
\subfigure[Multi-View Feature Selection\label{fig:Multi-view}]
{\includegraphics[width=\textwidth]{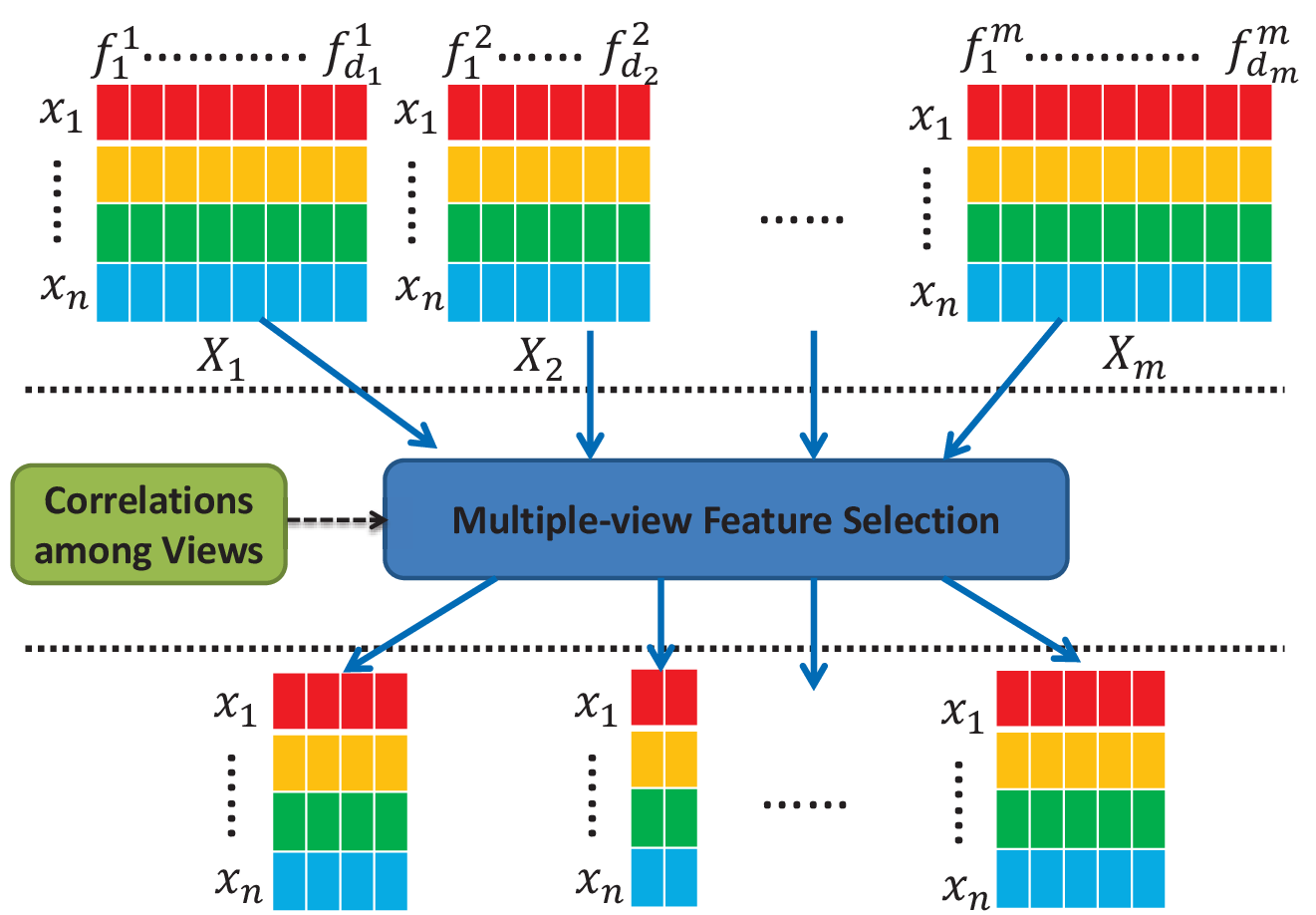}}
\end{minipage}
\caption{Differences between multi-source and multi-view feature selection.}
\vspace{-0.1in}
\label{fig:multi-source-multi-view}
\end{figure}

Multi-View data represent different facets of data instances in different feature spaces. These feature spaces are naturally dependent and high-dimensional. Hence, the task of multi-view feature selection arises~\cite{feng2013adaptive,tang2013unsupervised,wang2013multi,liu2016robust}, which aims to select features from different feature spaces simultaneously by using their relations. One motivating example is to select relevant features in pixels, tags, and terms associated with images simultaneously. Since multi-view feature selection is designed to select features across multiple views by using their relations, they are naturally different from multi-source feature selection. The difference between multi-source feature selection and multi-view feature selection is illustrated in Fig.~\ref{fig:multi-source-multi-view}.
For supervised multi-view feature selection, the most common approach is Sparse Group Lasso~\cite{friedman2010note,peng2010regularized}. In this subsection, we review some representative algorithms for unsupervised multi-view feature selection.

\vspace{-0.05in}
\subsubsection{Adaptive Multi-View Feature Selection}
Adaptive unsupervised multi-view feature selection (AUMFS)~\cite{feng2013adaptive} takes advantages of the data cluster structure, the data similarity and the correlations among views simultaneously. Specifically, let $\mat{X}_{1}\in\mathbb{R}^{n\times d_{1}},\mat{X}_{2}\in\mathbb{R}^{n\times d_{2}},...,\mat{X}_{1}\in\mathbb{R}^{n\times d_{m}}$ denote the description of $n$ instances from $m$ different views respectively, $\mat{X}=[\mat{X}_{1},\mat{X}_{2},...,\mat{X}_{m}]\in\mathbb{R}^{d}$ denotes the concatenated data, where $d=d_{1}+d_{2}+...+d_{m}$. AUMFS first builds a feature selection model by using $\ell_{2,1}$-norm regularized least squares loss function:
\begin{equation}
\small
\min_{\mat{W},\mat{F}}\|\mat{X}\mat{W}-\mat{F}\|_{2,1}+\alpha\|\mat{W}\|_{2,1},
\label{eq:AUMFS-obj1}
\vspace{-0.05in}
\end{equation}
where $\mat{F}\in\mathbb{R}^{n\times c}$ is the pseudo class label matrix. The $\ell_{2,1}$-norm loss function is imposed since it is robust to outliers and $\ell_{2,1}$-norm regularization selects features across all $c$ pseudo class labels with joint sparsity. Then AUMFS uses spectral clustering on an affinity matrix from different views to learn the shared pseudo class labels. For the data matrix $\mat{X}_{i}$ in each view, it first builds an affinity matrix $\mat{S}_{i}$ based on the data similarity on that view and gets the corresponding Laplacian matrix $\mat{L}_{i}$. Then it aims to learn the pseudo class label matrix by considering the spectral clustering from all views. Integrating it with Eq.~(\ref{eq:AUMFS-obj1}), we have the following objective function:
\begin{equation}
\small
\begin{split}
\min tr(\mat{F}'&\sum_{i=1}^{m}\lambda_{i}\mat{L}_{i}\mat{F})+\beta(\|\mat{X}\mat{W}-\mat{F}\|_{2,1}+\alpha\|\mat{W}\|_{2,1})\\
\mbox{s.t. }&\quad \mat{F}'\mat{F}=\mat{I}_{c}, \mat{F}\geq 0, \sum_{i=1}^{m}\lambda_{i}=1, \lambda_{i}\geq 0.
\end{split}
\label{eq:AUMFS-obj}
\vspace{-0.05in}
\end{equation}
where the contribution of each view for the joint spectral clustering is balanced by a nonnegative weight $\lambda_{i}$ and the summation of all $\lambda_{i}$ equals 1. $\beta$ is a parameter to balance the contribution of spectral clustering and feature selection.

\vspace{-0.05in}
\subsubsection{Unsupervised Feature Selection for Multi-View Data}
AUMFS~\cite{feng2013adaptive} learns one feature weight matrix for all features from different views to approximate the pseudo class labels. ~\cite{tang2013unsupervised} propose a novel unsupervised feature selection method called Multi-View Feature Selection (MVFS). Similar to AUMFS, MVFS uses spectral clustering with the affinity matrix from different views to learn the pseudo class labels. It differs from AUMFS as it learns one feature weight matrix for each view to fit the pseudo class labels by the joint least squares loss and $\ell_{2,1}$-norm regularization. The optimization problem of MVFS can be formulated as follows:
\begin{equation}
\small
\begin{split}
\min tr(\mat{F}'&\sum_{i=1}^{m}\lambda_{i}\mat{L}_{i}\mat{F})+\sum_{i=1}^{m}\beta(\|\mat{X}_{i}\mat{W}_{i}-\mat{F}\|_{2,1}+\alpha\|\mat{W}_{i}\|_{2,1})\\
\mbox{s.t. }&\quad \mat{F}'\mat{F}=\mat{I}_{c}, \mat{F}\geq 0, \sum_{i=1}^{m}\lambda_{i}=1, \lambda_{i}\geq 0.
\end{split}
\label{eq:MVFS-obj}
\vspace{-0.05in}
\end{equation}
The parameter $\lambda_{i}$ is employed to control the contribution of each view and $\sum_{i=1}^{m}\lambda_{i}=1$.

\vspace{-0.05in}
\subsubsection{Multi-View Clustering and Feature Learning via Structured Sparsity}
In some cases, features from a certain view contain more discriminative information than features from other views. One example is that in image processing, the color features are more useful than other types of features in identifying stop signs. To address this issue in multi-view feature selection, a novel feature selection algorithm is proposed in~\cite{wang2013multi} with a joint group $\ell_{1}$-norm and $\ell_{2,1}$-norm regularization.

For the feature weight matrix $\mat{W}_{1},...,\mat{W}_{m}$ from $m$ different views, the group $\ell_{1}$-norm is defined as $\|\mat{W}\|_{G_{1}}=\sum_{j=1}^{c}\sum_{i=1}^{m}\|\mat{W}_{i}(:,j)\|$. Crucially, the group $\ell_{1}$-norm regularization term is able to capture the global relations among different views and is able to achieve view-wise sparsity such that only a few views are selected. In addition to group $\ell_{1}$-norm, a $\ell_{2,1}$-norm regularizer on $\mat{W}$ is also included to achieve feature sparsity among selected views. Hence, the objective function of the proposed method is formulated as follows:
\begin{equation}
\small
\begin{split}
\min_{\mat{W},\mat{F}}&\|\mat{XW}-\mat{F}\|_{F}^{2}+\alpha\|\mat{W}\|_{2,1}+\beta\|\mat{W}\|_{G_{1}}\\
\mbox{s.t. }&\quad \mat{F}'\mat{F}=\mat{I}_{c}, \mat{F}\geq 0,
\end{split}
\end{equation}
where $\alpha$ and $\beta$ are used to control inter-view sparsity and intra-view sparsity.
\vspace{-0.02in}
\paragraph{\textbf{Discussion}: Feature selection algorithms for heterogeneous data can handle various types of data simultaneously. By fusing multiple data sources together, the selected features are able to capture the inherent characteristics of data and could better serve other learning problems on such data. However, most of the proposed algorithms in this family use matrices to represent the data and often convert the feature selection problem into an optimization algorithm. The resulted optimization problem often requires complex matrix operations which is computationally expensive, and also, limits the scalability of these algorithms for large-scale data. How to design efficient and distributed algorithms to speed up the computation is still a fertile area and needs deeper investigation}
\vspace{-0.05in}
\section{Feature Selection with Streaming Data}
Previous methods assume that all data instances and features are known in advance. However, it is not the case in many real-world applications that we are more likely faced with data streams and feature streams. In the worst cases, the size of data or features are unknown or even infinite. Thus it is not practical to wait until all data instances or features are available to perform feature selection. For streaming data, one motivating example online spam email detection problem, where new emails are continuously arriving; it is not easy to employ batch-mode feature selection methods to select relevant features in a timely manner. On an orthogonal setting, feature selection for streaming features also has its practical significances. For example, Twitter
produces more than 500 million tweets every day, and a large amount of slang words (features) are continuously being generated. These slang words promptly grab users' attention and become popular in a short time. Therefore, it is preferred to perform streaming feature selection to adapt to the changes on the fly. There are also some attempts to study these two dual problems together, which is referred as feature selection on Trapezoidal data streams~\cite{zhangtowards}. We will review some representative algorithms for these two orthogonal problems.
\vspace{-0.05in}
\subsection{Feature Selection Algorithms with Feature Streams}
For the feature selection problems with streaming features, the number of instances is considered to be constant while candidate features arrive one at a time; the task is to timely select a subset of relevant features from all features seen so far~\cite{perkins2003online,zhou2005streaming,wu2010online,yu2014towards,li2015unsupervised}. At each time step, a typical streaming feature selection algorithm first determines whether to accept the most recently arrived feature; if the feature is added to the selected feature set, it then determines whether to discard some existing features. The process repeats until no new features show up anymore. Different algorithms have different implementations in the first step. The second step which checks existing features is an optional step.
\vspace{-0.05in}
\subsubsection{Grafting Algorithm}
The first attempt to perform streaming feature selection is credited to~\cite{perkins2003online}. Their method is based on a stagewise gradient descent regularized risk framework~\cite{perkins2003grafting}. Grafting is a general technique that can deal with a variety of models that are parameterized by a feature weight vector $\mat{w}$ subject to $\ell_{1}$-norm regularization, such as Lasso.

The basic idea of Grafting is based on the observation -- incorporating a new feature into the Lasso model involves adding a new penalty term into the model. For example, at the time step $j$, when a new feature $f_{j}$ arrives, it incurs a regularization penalty of $\alpha|\mat{w}_{j}|$. Therefore, the addition of the new feature $f_{j}$ reduces the objective function value in Lasso only when the reduction in the loss function part $loss(\mat{w};\mat{X},\mat{y})$ outweighs the increase in the $\ell_{1}$-norm regularization. With this observation, the condition of accepting the new feature $f_{j}$ is $\left|\frac{\partial loss(\mat{w};\mat{X},\mat{y})}{\partial \mat{w}_{j}}\right|>\alpha$. Otherwise, the Grafting algorithm will set the feature coefficient $\mat{w}_{j}$ of the new feature $f_{j}$ to be zero. In the second step, when new features are accepted and included in the model, Grafting adopts a conjugate gradient (CG) procedure to optimize the model with respect to all current parameters to exclude some outdated features.
\vspace{-0.05in}
\subsubsection{Alpha-investing Algorithm}
Alpha-investing~\cite{zhou2005streaming} is an adaptive complexity penalty method which dynamically changes the threshold of error reduction that is required to accept a new feature. It is motivated by a desire to control the false discovery rate (FDR) of newly arrived features, such that a small portion of spurious features do not affect the model's accuracy significantly. The detailed algorithm works as follows: (1) it initializes $w_{0}=0$ (probability of false positives), $i=0$ (index of features), and selected features in the model to be empty; (2) it sets $\alpha_{i}=w_{i}/2i$ when a new feature arrives; (3) it sets $w_{i+1}=w_{i}-\alpha_{i}$ if $p\_value(f_{i},SF)\geq \alpha_{i}$; or set $w_{i+1}=w_{i}+\alpha_{\Delta}-\alpha_{i}$, $SF=SF\cup f_{i}$ if $p\_value(f_{i},SF)< \alpha_{i}$. The threshold $\alpha_{i}$ corresponds to the probability of selecting a spurious feature at the time step $i$. It is adjusted by the wealth $w_{i}$, which denotes the acceptable number of false positively detected features at the current moment. The wealth $w_{i}$ increases when a feature is added to the model. Otherwise, it decreases when a feature is not included to save for future features. More precisely, at each time step, the method calculates the $p$-value by using the fact that $\Delta$Logliklohood is equivalent to t-statistics. The $p$-value denotes the probability that a feature coefficient could be set to nonzero when it is not (false positively detected). The basic idea of alpha-investing is to adaptively adjust the threshold such that when new features are selected and included into the model, it allows a higher chance of including incorrect features in the future. On the other hand, each time when a new feature is not included, the wealth is wasted and lowers the chance of finding more spurious features.
\vspace{-0.05in}
\subsubsection{Online Streaming Feature Selection Algorithm}
Some other researchers study the streaming feature selection problem from an information theoretic perspective~\cite{wu2010online}. According to the definition, the whole feature set consists of four types of features: irrelevant, redundant, weakly relevant but non-redundant, and strongly relevant features. An optimal feature selection should select non-redundant and strongly relevant features. But as features continuously arrive in a streaming fashion, it is difficult to find all strongly relevant and non-redundant features. The proposed method, OSFS is able to capture these non-redundant and strongly relevant features via two steps: (1) online relevance analysis, and (2) online redundancy analysis. In the online relevance analysis step, OSFS discovers weakly relevant and strongly relevant features, and these features are added into the best candidate features (BCF). Otherwise, if the newly arrived feature is not relevant to the class label, it is discarded and not considered in future steps. In online redundancy analysis step, OSFS dynamically eliminates redundant features in the selected subset using Markov Blanket. For each feature $f_{j}$ in the best candidate set $BCF$, if there exists a subset of $BCF$ making $f_{j}$ and the class label conditionally independent, then $f_{j}$ is removed from $BCF$.
\vspace{-0.05in}
\subsubsection{Unsupervised Streaming Feature Selection in Social Media}
Vast majority of streaming feature selection methods are supervised which utilize label information to guide feature selection. However, in social media, it is easy to amass vast quantities of unlabeled data, while it is time and labor consuming to obtain labels. To deal with large-scale unlabeled data in social media, authors in~\cite{li2015unsupervised} propose the USFS algorithm to tackle unsupervised streaming feature selection. The key idea of USFS is to utilize source information such as link information. USFS first uncovers hidden social factors from link information by mixed membership stochastic blockmodel~\cite{airoldi2009mixed}. After obtaining the social latent factors $\mat{\Pi}\in\mathbb{R}^{n\times k}$ for each linked instance, USFS takes advantage of them as a constraint to perform selection. At a specific time step $t$, let $\mat{X}^{(t)}$, $\mat{W}^{(t)}$ denote the corresponding feature matrix and feature coefficient respectively. To model feature information, USFS constructs a graph $\mathcal{G}$ to represent feature similarity and $\mat{A}^{(t)}$ denotes the adjacency matrix of the graph, $\mat{L}^{(t)}$ is the corresponding Laplacian matrix from $\mat{X}^{(t)}$. Then the objective function to achieve feature selection at the time step $t$ is given as follows:
\begin{equation}
\small
\min_{\mat{W}^{(t)}}\frac{1}{2}\|\mat{X}^{(t)}\mat{W}^{(t)}-\boldsymbol{\Pi}\|_{F}^{2}+\alpha\sum_{i=1}^{k}\|(\mat{w}^{(t)})^{i}\|_{1}+\frac{\beta}{2}\|\mat{W}^{(t)}\|_{F}^{2}+\frac{\gamma}{2}\|(\mat{X}^{(t)}\mat{W}^{(t)})'(\mat{L}^{(t)})^{\frac{1}{2}}\|_{F}^{2},
\label{eq:USFS}
\vspace{-0.05in}
\end{equation}
where $\alpha$ is a sparse regularization parameter, $\beta$ controls the robustness of the model, and $\gamma$ balances link information and feature information. Assume at the next time step $t+1$ a new feature arrives, to test the new feature, USFS takes a similar strategy as Grafting to perform gradient test. Specifically, if the inclusion of the new feature is going to reduce the objective function in Eq.~(\ref{eq:USFS}), the feature is accepted; otherwise the new feature can be removed. When new features are continuously being generated, some existing features may become outdated, therefore, USFS also investigates if it is necessary to remove any existing features by re-optimizing the model through a BFGS method~\cite{boyd2004convex}.

\vspace{-0.05in}
\subsection{Feature Selection Algorithms with Data Streams}
In this subsection, we review the problem of feature selection with data streams which is considered as a dual problem of streaming feature selection.
\vspace{-0.05in}
\subsubsection{Online Feature Selection}
In~\cite{wang2014online}, an online feature selection algorithm (OFS) for binary classification is proposed. Let $\{\mat{x}_{1},\mat{x}_{2},...,\mat{x}_{t}...\}$ and $\{y_{1},y_{2},...,y_{t}...\}$ denote a sequence of input data instances and input class labels respectively, where each data instance $\mat{x}_{i}\in\mathbb{R}^{d}$ is in a $d$-dimensional space and class label $y_{i}\in\{-1,+1\}$. The task of OFS is to learn a linear classifier $\mat{w}^{(t)}\in\mathbb{R}^{d}$ that can be used to classify each instance $\mat{x}_{i}$ by a linear function sign(${\mat{w}^{(t)}}'\mat{x}_{i}$). To achieve feature selection, it requires that the linear classifier $\mat{w}^{(t)}$ has at most $B$-nonzero elements such that $\|\mat{w}^{(t)}\|_{0}\leq B$. It indicates that at most $B$ features will be used for classification. With a regularization parameter $\lambda$ and a step size $\eta$, the algorithm of OFS works as follows: (1) get a new data instance $\mat{x}_{t}$ and its class label $y_{t}$; (2) make a class label prediction sign(${\mat{w}^{(t)}}'\mat{x}_{t}$) for the new instance; (3) if $\mat{x}_{t}$ is misclassified such that $y_{i}{\mat{w}^{(t)}}'\mat{x}_{t}<0$, then $\tilde{\mat{w}}_{t+1}=(1-\lambda\eta)\mat{w}_{t}+\eta y_{t}\mat{x}_{t}$, $\hat{\mat{w}}_{t+1} = \min\{1,1/\sqrt{\lambda}\|\tilde{\mat{w}}_{t+1}\|_{2}\}\tilde{\mat{w}}_{t+1}$, and $\mat{w}_{t+1}=Truncate(\hat{\mat{w}}_{t+1},B)$; (4) $\mat{w}_{t+1}=(1-\lambda\eta)\mat{w}_{t}$. In particular, each time when a training instance $\mat{x}_{t}$ is misclassified, $\mat{w}_{t}$ is first updated by online gradient descent and then it is projected to a $\ell_{2}$-norm ball to ensure that the classifier is bounded. After that, the new classifier $\hat{\mat{w}}_{t+1}$ is truncated by taking the most important $B$ features. A subset of $B$ features is returned at each time step. The process repeats until there are no new data instances arrive anymore.
\vspace{-0.05in}
\subsubsection{Unsupervised Feature Selection on Data Streams}
To timely select a subset of relevant features when unlabeled data is continuously being generated,~\cite{huang2015unsupervised} propose a novel unsupervised feature selection method (FSDS) with only one pass of the data and with limited storage. The basic idea of FSDS is to use matrix sketching to efficiently maintain a low-rank approximation of the current observed data and then apply regularized regression to obtain the feature coefficients, which can further be used to obtain the importance of features. The authors empirically show that when some orthogonality conditions are satisfied, the ridge regression can replace the Lasso for feature selection, which is more computationally efficient. Assume at a specific time step $t$, $\mat{X}^{(t)}\in\mathbb{R}^{n_{t}\times d}$ denotes the data matrix at that time step, the feature coefficients can be obtained by minimizing the following:
\begin{equation}
\small
\min_{\mat{W}^{(t)}}\|\mat{B}^{(t)}\mat{W}^{(t)}-\{\mat{e}_{1},...,\mat{e}_{k}\}\|_{F}^{2}+\alpha\|\mat{W}^{(t)}\|_{F}^{2},
\label{eq:FSDS_obj2}
\end{equation}
where $\mat{B}^{(t)}\in\mathbb{R}^{\ell\times d}$ denote the sketching matrix of $\mat{X}^{(t)}$ ($\ell \ll n_{t}$), $\mat{e}_{i}\in\mathbb{R}^{\ell}$ is a vector with its $i$-th location as 1 and other locations as 0. By solving the optimization problem in Eq.~(\ref{eq:FSDS_obj2}), the importance of each feature $f_{i}$ is $score(j)=\max_{i}|\mat{W}^{(t)}(j,i)|$. The higher the feature score, the more important the feature is.
\vspace{-0.02in}
\paragraph{\textbf{Discussion}: As data is often not static and is generated in a streaming fashion, feature selection algorithms for both feature and data streams are often more desired in practical usage. Most of the existing algorithms in this family employ various strategies to speed up the selection process such that it can deal with new data samples or new features upon the arrival. However, it should be mentioned that most of these algorithms require multiple pass of the data and some even need to store all the historically generated data, which jeopardizes the usage of these algorithms when we only have limited memory or disk storage. It requires further efforts to design streaming algorithms that are effective and efficient with limited storage costs}
\vspace{-0.05in}
\section{Performance Evaluation}
We first introduce our efforts in developing an open-source feature selection repository. Then we use algorithms included in the repository as an example to show how to evaluate different feature selection algorithms.

\subsection{Feature Selection Repository}
First, we introduce our attempt in developing a feature selection repository -- \emph{scikit-feature}. The purpose of this feature selection repository is to collect some widely used feature selection algorithms that have been developed in the feature selection research to serve as a platform to facilitate their application, comparison, and joint study. The feature selection repository also effectively assists researchers to achieve more reliable evaluation in the process of developing new feature selection algorithms.

We develop the open source feature selection repository \emph{scikit-feature} by one of the most popular programming language -- python. It contains around 40 popular feature selection algorithms. It is built upon one widely used machine learning package \emph{scikit-learn} and two scientific computing packages \emph{Numpy} and \emph{Scipy}. At the same time, we also maintain a website (\url{http://featureselection.asu.edu/}) for this project which offers several sources such as publically available benchmark datasets, performance evaluation of algorithms and test cases to run each algorithm. The source code of this repository is available at Github (\url{https://github.com/jundongl/scikit-feature}). An interactive tool of the repository is also available~\cite{cheng2016featureminer}. We welcome researchers in this community to contribute algorithms and datasets to our repository.

\subsection{Evaluation Methods and Metrics}
As an example, we empirically show how to evaluate the performance of feature selection algorithms in the repository. The experimental results can be obtained from our repository project website (\url{http://featureselection.asu.edu/datasets.php}). In our project website, for each dataset, we list all applicable feature selection algorithms along with its evaluation on either classification or clustering. Next, we will provide detailed information how these algorithms are evaluated, including evaluation criteria and experimental setup. Different feature selection algorithms can be categorized by the following two criteria: (1) labels: supervised or unsupervised; (2) output: feature weighting or subset selection. The first criterion determines whether we need to use the label information to perform feature selection or not. The second criterion categorizes these algorithms based on the output. Feature weighing algorithms give each feature a score for ranking and feature subset algorithms only show which features are selected.

Next, we introduce the widely adopted way to evaluate the performance of feature selection algorithms. We have different evaluation metrics for supervised and unsupervised methods. For different output types, different evaluation strategies are used: (1) if it is a feature weighting method that outputs the feature scores, then the quality of the first $\{5,10,15,...,295,300\}$ features are evaluated respectively; (2) if it is a feature subset selection method that only outputs which features are selected, then we use all the selected features to perform the evaluation.
\vspace{-0.05in}
\paragraph{Supervised Methods}
To test the performance of supervised feature selection algorithms, we divide the whole dataset into two parts - the training set $\mathcal{T}$ and test set $\mathcal{U}$. Feature selection algorithms will be first applied to the training set $\mathcal{T}$ to obtain a subset of relevant features $\mathcal{S}$. Then the test set on the selected features acts as input to a classification model for the testing purpose. In the experiments, we use classification accuracy to evaluate the classification performance and three classification models, Linear SVM, Decision Tree, Na\"{\i}ve Bayes are used. To get more reliable results, 10-folds cross validation is used. Normally, the higher the classification performance, the better the selected features are.
\vspace{-0.05in}
\paragraph{Unsupervised Methods}
Following the standard way to assess unsupervised feature selection, we evaluate unsupervised algorithms in terms of clustering performance. Two commonly used clustering performance metrics~\cite{cai2010unsupervised}, i.e., \emph{normalized mutual information} (NMI) and \emph{accuracy} (ACC) are used. Each feature selection algorithm is first applied to select features; then K-means clustering is performed based on the selected features. We repeat the K-means algorithm 20 times and report the average clustering results since K-means may converge to a local optimal. The higher the clustering performance, the better the selected features are.

We also list the following information of main algorithms reviewed in this paper in Table~\ref{table:details}: (1) the type of data: conventional data or other types of data; (2) usage of labels: supervised or unsupervised\footnote{feature selection for regression can also be regarded as a supervised method, here we focus on feature selection for classification problems}; (3) output: feature weighting or subset selection; (4) feature type: numerical variables or discrete variables (numerical variables can also be divided into continuous variables and discrete variables). For supervised feature selection methods, we also list if the methods are designed to tackle binary-class or multi-class classification problems. Based on the above information, the practitioners can have a more intuitive sense about the applicable scenarios of different methods.
\begin{sidewaystable}[!htbp]
\centering
\caption{Detailed information of main feature selection algorithms reviewed in the paper.}
\label{table:details}
\resizebox{\textwidth}{!}{%
\begin{tabular}{|c|c|c|c|c|c|c|c|c|c|}
\hline
\multirow{3}{*}{Data} & \multirow{3}{*}{Methods} & \multicolumn{3}{c|}{Supervision} & \multicolumn{2}{c|}{Output of Features} & \multicolumn{3}{c|}{Feature Type} \\ \cline{3-10}
 &  & \multirow{2}{*}{Binary} & \multirow{2}{*}{Multi-class} & \multirow{2}{*}{Unsupervised} & \multirow{2}{*}{Ranking} & \multirow{2}{*}{Subset} & \multicolumn{2}{c|}{Numerical} & \multirow{2}{*}{Categorical} \\ \cline{8-9}
 &  &  &  &  &  &  & Continuous & Discrete &  \\ \hline
\multirow{27}{*}{Conventional--Flat Features} & Fisher Score~\cite{duda2012pattern} & \checkmark & \checkmark &  & \checkmark &  & \checkmark & \checkmark &  \\ \cline{2-10}
 & ReliefF~\cite{robnik2003theoretical} & \checkmark & \checkmark &  & \checkmark &  & \checkmark & \checkmark &  \\ \cline{2-10}
 & Trace Ratio~\cite{nie2008trace} & \checkmark & \checkmark &  & \checkmark &  & \checkmark & \checkmark &  \\ \cline{2-10}
 & Laplacian Score~\cite{he2005laplacian} &  &  & \checkmark & \checkmark &  & \checkmark & \checkmark &  \\ \cline{2-10}
 & SPEC~\cite{zhao2007spectral} & \checkmark & \checkmark & \checkmark & \checkmark &  & \checkmark & \checkmark &  \\ \cline{2-10}
 & MIM~\cite{lewis1992feature} & \checkmark & \checkmark &  & \checkmark &  &  & \checkmark & \checkmark \\ \cline{2-10}
 & MIFS~\cite{battiti1994using} & \checkmark & \checkmark &  & \checkmark &  &  & \checkmark & \checkmark \\ \cline{2-10}
 & MRMR~\cite{peng2005feature} & \checkmark & \checkmark &  & \checkmark &  &  & \checkmark & \checkmark \\ \cline{2-10}
 & CIFE~\cite{lin2006conditional} & \checkmark & \checkmark &  & \checkmark &  &  & \checkmark & \checkmark \\ \cline{2-10}
 & JMI~\cite{meyer2008information} & \checkmark & \checkmark &  & \checkmark &  &  & \checkmark & \checkmark \\ \cline{2-10}
 & CMIM~\cite{fleuret2004fast} & \checkmark & \checkmark &  & \checkmark &  &  & \checkmark & \checkmark \\ \cline{2-10}
 & IF~\cite{vidal2003object} & \checkmark & \checkmark &  & \checkmark &  &  & \checkmark & \checkmark \\ \cline{2-10}
 & ICAP~\cite{jakulin2005machine} & \checkmark & \checkmark &  & \checkmark &  &  & \checkmark & \checkmark \\ \cline{2-10}
 & DISR~\cite{meyer2006use} & \checkmark & \checkmark &  & \checkmark &  &  & \checkmark & \checkmark \\ \cline{2-10}
 & FCBF~\cite{yu2003feature} & \checkmark & \checkmark &  &  & \checkmark &  & \checkmark & \checkmark \\ \cline{2-10}
 & $\ell_{p}$-regularized~\cite{Liu:2009:SLEP:manual} & \checkmark &  &  & \checkmark &  & \checkmark & \checkmark &  \\ \cline{2-10}
 & $\ell_{p,q}$-regularized~\cite{Liu:2009:SLEP:manual} &  & \checkmark &  & \checkmark &  & \checkmark & \checkmark &  \\ \cline{2-10}
 & REFS~\cite{nie2010efficient} & \checkmark & \checkmark &  & \checkmark &  & \checkmark & \checkmark &  \\ \cline{2-10}
 & MCFS~\cite{cai2010unsupervised} &  &  & \checkmark & \checkmark &  & \checkmark & \checkmark &  \\ \cline{2-10}
 & UDFS~\cite{yang2011l2} &  &  & \checkmark & \checkmark &  & \checkmark & \checkmark &  \\ \cline{2-10}
 & NDFS~\cite{li2012unsupervised} &  &  & \checkmark & \checkmark &  & \checkmark & \checkmark &  \\ \cline{2-10}
 & Low Variance~\cite{pedregosa2011scikit} &  &  & \checkmark &  & \checkmark &  & \checkmark & \checkmark \\ \cline{2-10}
 & T-score~\cite{davis1986statistics} & \checkmark &  &  & \checkmark &  & \checkmark & \checkmark &  \\ \cline{2-10}
 & Chi-square~\cite{liu1995chi2} & \checkmark & \checkmark &  & \checkmark &  &  & \checkmark & \checkmark \\ \cline{2-10}
 & Gini~\cite{gini1912variability}& \checkmark & \checkmark &  & \checkmark &  &  & \checkmark & \checkmark \\ \cline{2-10}
 & CFS~\cite{hall1999feature} & \checkmark & \checkmark &  &  & \checkmark &  & \checkmark & \checkmark \\ \hline
\multirow{6}{*}{Conventional -- Structured Feature} & Group Lasso~\cite{} & \checkmark &  &  & \checkmark &  & \checkmark & \checkmark &  \\ \cline{2-10}
 & Sparse Group Lasso~\cite{friedman2010note} & \checkmark &  &  & \checkmark &  & \checkmark & \checkmark &  \\ \cline{2-10}
 & Tree Lasso~\cite{liu2010moreau} & \checkmark &  &  & \checkmark &  & \checkmark & \checkmark &  \\ \cline{2-10}
 & Graph Lasso~\cite{ye2012sparse} & \checkmark &  &  & \checkmark &  & \checkmark & \checkmark &  \\ \cline{2-10}
 & GFLasso~\cite{kim2009statistical} & \checkmark &  &  & \checkmark &  & \checkmark & \checkmark &  \\ \cline{2-10}
 & GOSCAR~\cite{yang2012feature} & \checkmark &  &  & \checkmark &  & \checkmark & \checkmark &  \\ \hline
\multirow{4}{*}{Linked Data} & FSNet~\cite{gu2011towards} & \checkmark & \checkmark &  & \checkmark &  & \checkmark & \checkmark &  \\ \cline{2-10}
 & LinkedFS~\cite{tang2012feature} & \checkmark & \checkmark &  & \checkmark &  & \checkmark & \checkmark &  \\ \cline{2-10}
 & LUFS~\cite{tang2012unsupervised} &  &  & \checkmark & \checkmark &  & \checkmark & \checkmark &  \\ \cline{2-10}
 & NetFS~\cite{li2016robust} &  &  & \checkmark & \checkmark &  & \checkmark & \checkmark &  \\ \hline
Multi-Source & GDCOV~\cite{zhao2008multi} &  &  & \checkmark & \checkmark &  & \checkmark & \checkmark &  \\ \hline
\multirow{2}{*}{Multi-View} & AUMFS~\cite{feng2013adaptive} &  &  & \checkmark & \checkmark &  & \checkmark & \checkmark &  \\ \cline{2-10}
 & MVFS~\cite{tang2013unsupervised} &  &  & \checkmark & \checkmark &  & \checkmark & \checkmark &  \\ \hline
\multirow{4}{*}{Streaming Feature} & Grafting~\cite{perkins2003online} & \checkmark &  &  &  & \checkmark & \checkmark & \checkmark &  \\ \cline{2-10}
 & Alpha-Investing~\cite{zhou2005streaming} & \checkmark &  &  &  & \checkmark & \checkmark & \checkmark &  \\ \cline{2-10}
 & OSFS~\cite{wu2010online} & \checkmark &  &  &  & \checkmark &  & \checkmark & \checkmark \\ \cline{2-10}
 & USFS~\cite{li2015unsupervised} &  &  & \checkmark &  & \checkmark & \checkmark & \checkmark &  \\ \hline
\multirow{2}{*}{Streaming Data} & OFS~\cite{wang2014online} & \checkmark &  &  & \checkmark &  & \checkmark & \checkmark &  \\ \cline{2-10}
 & FSDS~\cite{huang2015unsupervised} &  &  & \checkmark & \checkmark &  & \checkmark & \checkmark &  \\ \hline
\end{tabular}%
}
\end{sidewaystable}

\vspace{-0.1in}
\section{Open Problems and Challenges}
Over the past two decades, there has been a significant number of attempts in developing feature selection algorithms for both theoretical analysis and real-world applications. However, we still believe there is more work that can be done in this field. Here are several challenges and concerns that we need to mention and discuss.
\vspace{-0.05in}
\subsection{Scalability}
With the tremendous growth in the size of data, the scalability of most current feature selection algorithms may be jeopardized. In many scientific and business applications, data is usually measured in terabyte (1TB = $10^{12}$ bytes). Normally, datasets in the scale of terabytes cannot be loaded into the memory directly and therefore limits the usability of most feature selection algorithms. Currently, there are some attempts to use distributed programming frameworks to perform parallel feature selection for large-scale datasets~\cite{singh2009parallel,zhao2013massively,yamada2014n,zadeh2017scalable}. In addition, most of the existing feature selection methods have a time complexity proportional to $O(d^{2})$ or even $O(d)^{3}$, where $d$ is the feature dimension. Recently, big data of ultrahigh-dimensionality has emerged in many real-world applications such as text mining and information retrieval. Most feature selection algorithms do not scale well on the ultrahigh-dimensional data whose efficiency deteriorates quickly or is even computationally infeasible. In this case, well-designed feature selection algorithms in linear or sublinear running time are preferred~\cite{fan2009ultrahigh,tan2014towards}. Moreover, in some online classification or online clustering tasks, the scalability of feature selection algorithms is also a big issue. For example, the data streams or feature streams may be infinite and cannot be loaded into the memory, hence we can only make one pass of the data where the second pass is either unavailable or computationally expensive. Even though feature selection algorithms can reduce the issue of scalability for online classification or clustering, these methods either require to keep all features in the memory or require iterative processes to visit data instances more than once, which limit their practical usage. In conclusion, even though there is some preliminary work to increase the scalability of feature selection algorithms, we believe that more focus should be given to the scalability problem to keeping pace with the rapid growth of very large-scale and streaming data.
\vspace{-0.05in}
\subsection{Stability}
For supervised feature selection algorithms, their performance is usually evaluated by the classification accuracy. In addition to accuracy, the stability of these algorithms is also an important consideration when developing new feature selection algorithms. It is defined as the sensitivity of a feature selection algorithm to perturbation in the training data~\cite{kalousis2007stability,he2010stable,saeys2008robust,loscalzo2009consensus,yang2011robust}. The perturbation of data could be in various format such as addition/deletion of data samples and the inclusion of noisy/outlier samples. More rigorous definition on the stability of feature selection algorithms can be referred to~\cite{kalousis2007stability}. The stability of feature selection algorithms has significant implications in practice as it can help domain experts gain more confidence on the selected features. A motivating example in bioinformatics indicates that domain experts would like to see the same set or similar set of genes (features) selected each time when they obtain new data samples. Otherwise, domain experts would not trust these algorithms and may never use them again. It is observed that many well-known feature selection algorithms suffer from the low stability problem after the small data perturbation is introduced in the training set. It is also found in~\cite{alelyani2011effect} that the underlying characteristics of data may greatly affect the stability of feature selection algorithms and the stability issue may also be data dependent. These factors include the dimensionality of the feature, the number of data instances, etc. In against with supervised feature selection, the stability of unsupervised feature selection algorithms has not been well studied yet. Studying stability for unsupervised feature selection is much more difficult than that of the supervised methods. The reason is that in
unsupervised feature selection, we do not have enough prior knowledge about the cluster structure of the data. Thus, we are uncertain that if the new data instance, i.e., the perturbation belongs to any existing clusters or will introduce new clusters. While in supervised feature selection, we have prior knowledge about the label of each data instance, and a new sample that does not belong to any existing classes will be considered as an outlier and we do not need to modify the selected feature set to adapt to the outliers. In other words, unsupervised feature selection is more sensitive to noise and the noise will affect the stability of these algorithms.
\vspace{-0.05in}
\subsection{Model Selection}
For most feature selection algorithms especially for feature weighting methods, we have to specify the number of selected features. However, it is often unknown what is the optimal number of selected features. A large number of selected features will increase the risk in including noisy, redundant and irrelevant features, which may jeopardize the learning performance. On the other hand, it is also not good to include too small number of selected features, since some relevant features may be eliminated. In practice, we usually adopt a heuristic way to grid search the number of selected features and pick the number that has the best classification or clustering performance, but the whole process is computationally expensive. It is still an open and challenging problem to determine the optimal number of selected features. In addition to the optimal number of selected features, we also need to specify the number of clusters or pseudo classes for unsupervised feature selection algorithms. In real-world problems, we usually have limited knowledge about the clustering structure of the data. Choosing different numbers of clusters may merge totally different small clusters into one big cluster or split one big cluster into smaller ones. As a consequence, it may result in finding totally different subsets of features. Some work has been done to estimate these tricky parameters. For instance, in~\cite{tibshirani2001estimating}, a principled way to estimate the number of suitable clusters in a dataset is proposed. However, it is still not clear how to find the best number of clusters directly for unsupervised feature selection. All in all, we believe that the model selection is an important issue and needs deeper investigation.

\vspace{-0.05in}
\section{Conclusion}
Feature selection is effective in preprocessing data and reducing data dimensionality. Meanwhile, it is essential to successful data mining and machine learning applications. It has been a challenging research topic with practical significance in many areas such as statistics, pattern recognition, machine learning, and data mining (including web, text, image, and microarrays). The objectives of feature selection include: building simpler and more comprehensive models, improving data mining performance, and helping prepare clean and understandable data. The past few years have witnessed the development of many new feature selection methods. This survey article aims to provide a comprehensive review about recent advances in feature selection. We first introduce basic concepts of feature selection and emphasize the importance of applying feature selection algorithms to solve practical problems. Then we classify conventional feature selection methods from the label perspective and the selection strategy perspective. As current categorization cannot meet the rapid development of feature selection research especially in the era of big data, we propose to review recent advances in feature selection algorithms from a data perspective. In particular, we survey feature selection algorithms in four parts: (1) feature selection with conventional data with flat features; (2) feature selection with structured features; (3) feature selection with heterogeneous data; and (4) feature selection with streaming data. Specifically, we further classify conventional feature selection algorithms for conventional data (flat features) into similarity based, information theoretical based, sparse learning based and statistical based methods, and other types of methods according to the used techniques. For feature selection with structured features, we consider three types of structured features, namely group, tree and graph features. The third part studies feature selection with heterogeneous data, including feature selection with linked data, multi-source and multi-view feature selection. The last part consists of feature selection algorithms for streaming data and streaming features. We analyze the advantages and shortcomings of these different types of feature selection algorithms. To facilitate the research on feature selection, this survey is accompanied by a feature selection repository - \emph{scikit-feature}, which includes some of the most popular feature selection algorithms that have been developed in the past few decades. Some suggestions are given on how to evaluate these feature selection algorithms, either supervised or unsupervised methods. At the end of the survey, we present some open problems that require future research. It also should be mentioned that the aim of the survey is not to claim the superiority of some feature selection algorithms over others, but to provide a comprehensive structured list of recent advances in feature selection algorithms from a data perspective and a feature selection repository to promote the research in this community.


\end{document}